\pdfoutput=1

\documentclass[11pt]{article}

\usepackage{hyperref}

\usepackage{EMNLP2023}

\usepackage{times}
\usepackage{latexsym}

\usepackage{xcolor}
\usepackage{booktabs}
\usepackage{titlesec}
\usepackage{todonotes}

\usepackage{url}
\usepackage{lipsum}

\usepackage{bm}
\usepackage{physics}
\usepackage{bbm}
\usepackage{amsmath}
\usepackage{amsfonts,amssymb}

\usepackage{multirow}

\usepackage[T1]{fontenc}

\usepackage[utf8]{inputenc}

\usepackage{microtype}
\usepackage{mathtools}
\usepackage{nicefrac}
\usepackage{enumitem}
\usepackage{graphicx}
\usepackage[nameinlink,capitalize,noabbrev]{cleveref}
\usepackage{subcaption}

\usepackage{inconsolata}

\title{Mean BERTs make erratic language teachers: \\ the effectiveness of latent bootstrapping in low-resource settings}

\author{David Samuel \\
  University of Oslo, Language Technology Group \\
}

\begin{document}
\maketitle
\begin{abstract}
This paper explores the use of latent bootstrapping, an alternative self-supervision technique, for pretraining language models. Unlike the typical practice of using self-supervision on discrete subwords, latent bootstrapping leverages contextualized embeddings for a richer supervision signal. We conduct experiments to assess how effective this approach is for acquiring linguistic knowledge from limited resources. Specifically, our experiments are based on the BabyLM shared task, which includes pretraining on two small curated corpora and an evaluation on four linguistic benchmarks.
\end{abstract}

\section{Introduction}

All modern language models are trained with a general self-supervised learning (SSL)  paradigm \citep{Radford2018ImprovingLU, devlin-etal-2019-bert, 2020t5}. Recently, the field of visual representation learning has seen a growing usage of self-supervision on \textit{latent embeddings} \citep{10.5555/3495724.3497510, 10.5555/3524938.3525087, chen2020simsiam, assran2023self}. 
While this type of self-supervision has been recently proposed as an integral part of a human-like machine intelligence system \citep{lecun2022path}, language models are still mostly self-supervised on hard targets, typically on subword tokens.

The concept of \textit{latent bootstrapping} \citep{10.5555/3495724.3497510} offers a promising alternative, as the latent vectors provide a deep and semantically rich representation of the input. This, in turn, delivers a more valuable supervision signal compared to the conventional method of supervision on discrete subword indices. 
Data2vec \citep{pmlr-v162-baevski22a} showed that latent bootstrapping performs on par with traditional self-supervised language modeling when pretrained on a large text corpus. We argue that, intuitively, the rich training signal from contextualized embeddings should be particularly effective in low-resource data settings.

In this paper, our aim is to test this hypothesis and identify possible drawbacks of the bootstrapping method. We base our experiments on the \textit{BabyLM challenge} \citep{warstadt-et-al-2023-babylm}, a shared task that uses two carefully curated, sample-efficient pretraining corpora, mimicking the English language exposure to young children. In addition, this challenge employs four benchmarks to evaluate different aspects of linguistic knowledge and understanding learned by language models.

\begin{figure}[!t]
    \includegraphics[width=\columnwidth]{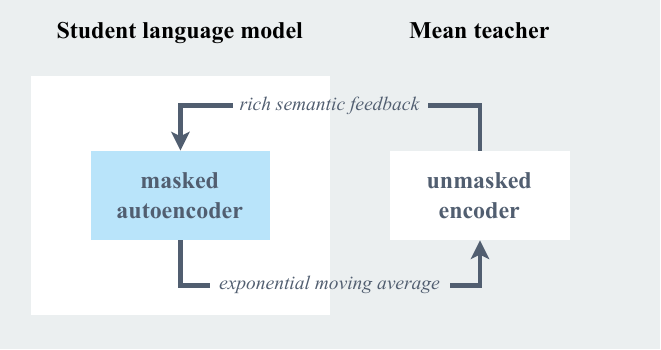}
    \caption{
    The self-supervision feedback loop of latent bootstrapping: a student model improves by aligning with its teacher's latent outputs and the teacher improves by maintaining the exponential moving average of the student.}
    \label{fig:loop}
\end{figure}

We introduce BootBERT, a novel masked autoencoder language model \citep{meng2023representation} that harnesses latent bootstrapping \citep{10.5555/3495724.3497510} between a mean teacher \citep{10.5555/3294771.3294885} and its student. Through a positive feedback loop, the student and the teacher iteratively learn from each other, as illustrated in \cref{fig:loop}. The student is trained to match the teacher’s outputs while the \emph{mean} teacher is defined as the exponential moving average of the student. Once pretraining is complete, only the student language model is used for evaluation and the teacher is discarded. We assess its performance on the BabyLM challenge, contrasting it with conventional language models. The source code and pretrained models are available online at {\footnotesize{\url{https://github.com/ltgoslo/boot-bert}}}.

\begin{figure*}[!th]
    \includegraphics[width=\textwidth]{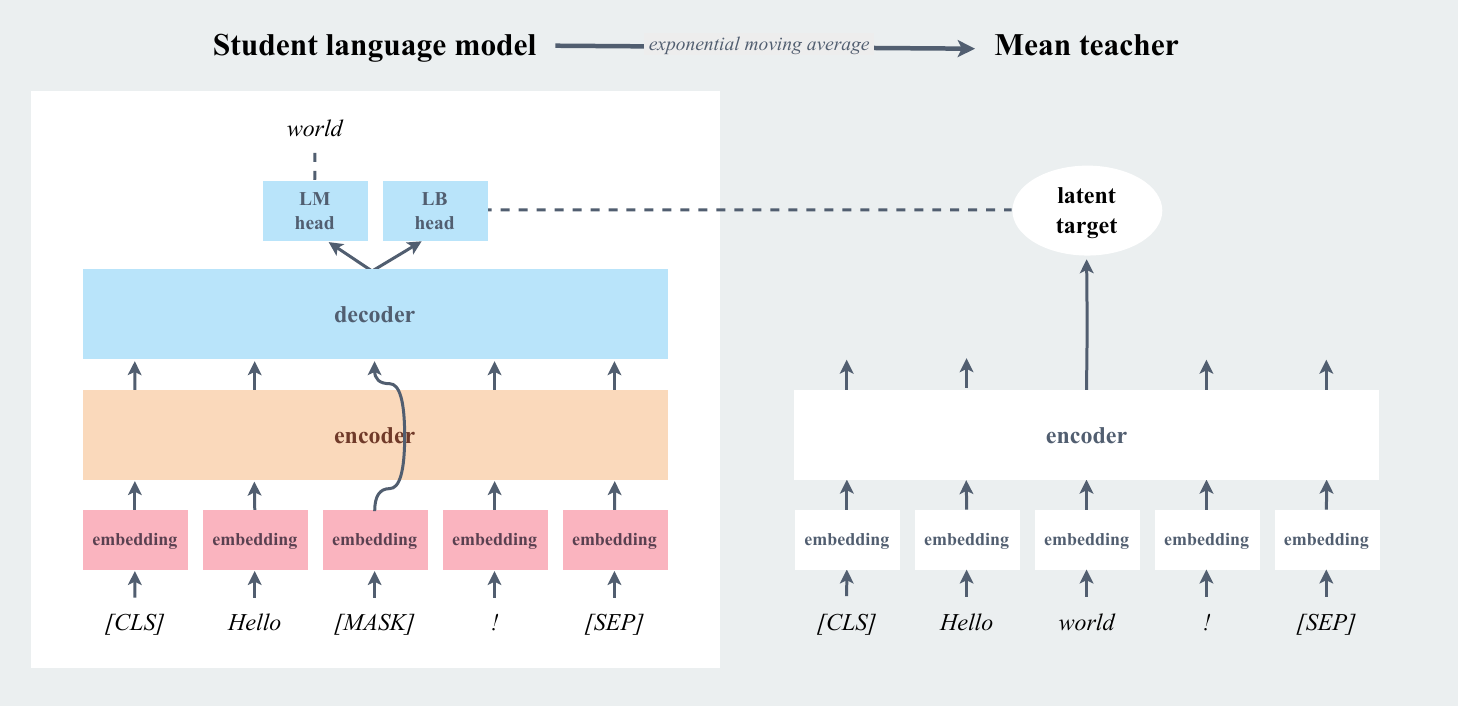}
    \caption{A detailed overview of the self-supervised feedback loop. The left side illustrates the student language model, a masked autoencoder network, that targets two training objectives: 1) conventional masked language modeling, aiming to predict the masked token (e.g., the word `world'), and 2) aligning the contextualized embedding of the masked tokens to their unmasked counterparts. The embeddings for the unmasked tokens are produced by a mean teacher network (on the right), computed as an exponential moving average of the student parameters.}
    \label{fig:encoder-decoder}
\end{figure*}

\section{Method}
\label{sec:method}

In this section, we outline our proposed model, BootBERT, delving into its neural architecture and the latent bootstrapping training objective. In order to allow for language-modeling-based evaluation, the bootstrapping objective operates alongside conventional masked language modeling. The diagram in \cref{fig:encoder-decoder} illustrates the general idea of this approach.

\paragraph{Masked autoencoder architecture.} BootBERT diverges slightly from the standard `encoder-only' architecture often found in masked language models \citep{devlin-etal-2019-bert}.  Instead, following the method of \newcite{meng2023representation}, we employ a masked autoencoder \citep[MAE;][]{He_2022_CVPR} framework for the text domain. This approach distinguishes the \textit{encoding} of contextualized embeddings from the \textit{decoding} of masked subwords. These two functionalities are separated by dividing the model into an encoder and a decoder module, as illustrated in \cref{fig:encoder-decoder} on the left.

The encoder's role is to create a bidirectional contextualized embedding of input tokens. Unlike traditional masked language models, the encoder does not process any \texttt{[MASK]} tokens, thus eliminating the need to allocate parameters for representing them \citep{meng2023representation}.

The \texttt{[MASK]} tokens are processed and denoised by the decoder module. The decoder is supplied with the full input -- the unmasked tokens are represented by their contextualized embeddings (provided by the encoder) and the masked tokens are represented by a static \texttt{[MASK]} embedding. Note that the decoder in this type of model is bidirectional and purely self-attentive, differing from the original definition of a transformer decoder by \newcite{NIPS2017_3f5ee243}.

\paragraph{Teacher-student feedback loop.} Conceptually, the training process can be divided into optimization of a student model and optimization of a teacher model. Here, the masked student autoencoder model is trained to match the contextualized embeddings of the \textit{unmasked} tokens, produced by the mean teacher network. In line with \newcite{10.5555/3294771.3294885}, the teacher parameters $\phi$ are not optimized via gradient descent, but rather through a slow exponential moving average (EMA) of the student parameters $\theta$:
\begin{equation*}
\phi = \tau\phi + (1-\tau)\theta.
\end{equation*}

\noindent
This moving average not only stabilizes the latent targets but also prevents representation collapse \citep{10.5555/3495724.3497510}.

\paragraph{Loss.} We optimize two objectives during training the student model: a traditional masked language modeling objective with hard targets, symbolized by $\mathcal{L}_{\mathrm{LM}}$, and a latent bootstrapping objective using teacher's latent targets $\mathcal{L}_{\mathrm{LB}}$. The final loss function combines these objectives with a weighted sum:
\begin{equation*}
\mathcal{L} = \mathcal{L}_{\mathrm{LB}} + \beta\mathcal{L}_{\mathrm{LM}}.
\end{equation*}

\noindent
Here, $\mathcal{L}_{\mathrm{LM}}$ is calculated simply as negative log-likelihood of the true targets. Its purpose is two-fold: allowing for a MLM-based evaluation (for example BLiMP), and preventing representation collapse of unconstrained latent bootstrapping \citep{10.5555/3495724.3497510}.

The second objective is computed as a smooth L1 loss between student predictions $y_s$ and teacher's contextualized embeddings $y_t$, This works mostly like a standard mean-squared error but prevents exploding gradients from outliers \citep{10.1109/ICCV.2015.169}:
\begin{equation*}
\mathcal{L}_\text{LB}(y_t, y_s) = 
\begin{cases}
0.5(y_t - y_s)^2 & | y_t - y_s | \leq 1 \\
|y_t-y_s|-0.5 & \text{otherwise.}
\end{cases}
\nonumber
\end{equation*}

\begin{figure}[!t]
    \includegraphics[width=\columnwidth]{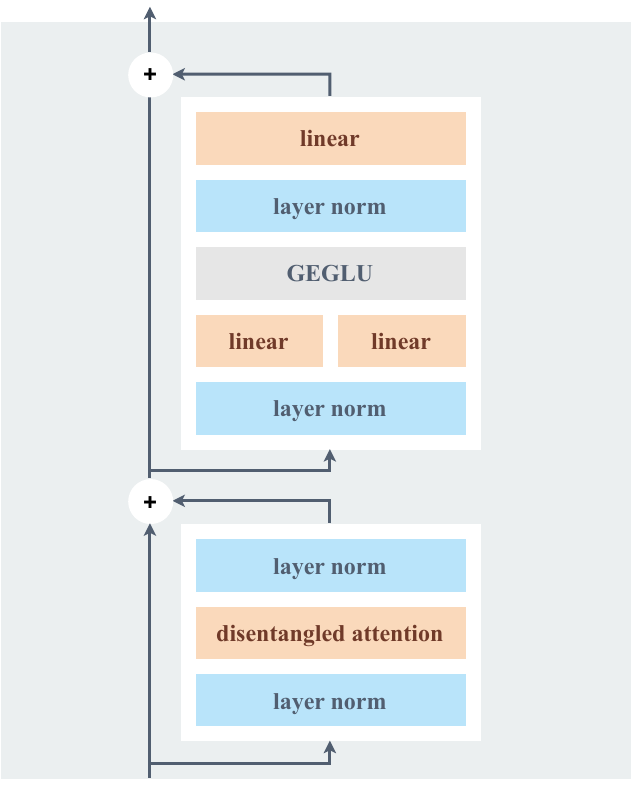}
    \caption{We base our model on LTG-BERT. This simplified diagram shows one layer from that transformer architecture, it illustrates the self-attention module (bottom) and the feed-forward module (top). Both modules utilize a modified NormFormer-like layer normalization placement and the feed-forward module contains a gated-linear activation function.}
    \label{fig:ltg-bert}
\end{figure}

\paragraph{LTG-BERT transformer backbone.} As for more low-level architectural and training choices, we adopt the approach of LTG-BERT by \newcite{samuel-etal-2023-trained}. This method was optimized for low-resource masked language modeling on a similar corpus to the corpora provided in BabyLM. The key improvements of the LTG-BERT transformer architecture include the use of the NormFormer layer normalization
\citep{shleifer2022normformer}, an alternative disentangled attention mechanism with relative positions \citep{he2021deberta} and gated-linear activation function \citep[GEGLU;][]{DBLP:journals/corr/abs-2002-05202}; as illustrated in \cref{fig:ltg-bert}. On top of these architectural changes, the authors also employ masking of random subword spans \citep{joshi-etal-2020-spanbert}. More details about these choices can be found in \newcite{samuel-etal-2023-trained}.

\section{Experiments}

The main goal of this paper is to evaluate how well language models trained with latent bootstrapping acquire language and if it makes a viable training objective for language representation learning. We base the experiments on the BabyLM challenge \citep{warstadt-et-al-2023-babylm}. First, we describe the pretraining process of two BabyLM tracks and second, the evaluation of pretrained models using the BabyLM evaluation pipeline.

\paragraph{BabyLM challenge.}

This challenge provides a share ground for experiments on small-scale language modeling. It consists of three tracks: \textsc{strict}, \textsc{strict-small} and \textsc{loose}. For the first two tracks, the submissions have to be pretrained solely on a fixed corpus provided by the organizers. This corpus contains about 100M words in the \textsc{strict} track and about 10M words in the \textsc{strict-small} track. As for the \textsc{loose} track, the submissions are still limited to pretrained on 100M words, but this data can come from any source and the models can utilize an unlimited amount of non-linguistic data in addition. As detailed in \cref{sec:evaluation}, the submissions are compared on a shared evaluation set consisting of syntactic and natural language understanding tasks.

\subsection{Pretraining}

The pretraining is done on corpora provided by the BabyLM challenge. These texts are curated specifically to be of the same type and quantity that children learn from. Thus, it allows us to assess (to some degree) whether latent bootstrapping is a more plausible cognitive model of human language acquisition \citep{warstadt-et-al-2023-babylm}.

\paragraph{Training corpus.}

Specifically, we consider the \textsc{strict} and \textsc{strict-small} tracks and pretrain the models on their respective 100-million-word and 10-million-word corpora. Both datasets contain child-directed speech, transcribed speech, children’s books and Wikipedia, among other sources. The content of these datasets is detailed in \cref{app:data-preprocess}, together with our simple preprocessing pipeline, which unifies the typographical features of the BabyLM subcorpora.

\paragraph{Pretraining process.} Generally speaking, we adopt the training recipe of LTG-BERT \citep{samuel-etal-2023-trained}, which was optimized for pretraining on another low-resource 100 million English corpus. The pretraining process is the same for both tracks, except for using a smaller vocabulary and a smaller model for the \textsc{strict-small} track.

As for the \textsc{strict} track, we use a \textsc{base}-size language model -- 12 encoder layers and 4 decoder layers with hidden size of 768 and with 12 attention heads. We train a case-sensitive WordPiece tokenizer \citep{https://doi.org/10.48550/arxiv.1609.08144} with a vocabulary size of $2^{14} = 16\,384$, using solely texts from the \textsc{strict} corpus. As per \newcite{samuel-etal-2023-trained}, we pretrain the models with \nicefrac{1}{2} of the \textsc{BERT} training budget, as it has been shown to be sufficient for a relatively small 100-million-word corpus. The tokens are masked with continuous span masking \citep{joshi-etal-2020-spanbert, 2020t5}. In particular, the masks are iteratively sampled until 15\% of tokens are masked and the length of each span is sampled from the geometric distribution $\textrm{Geo}(p)$, with $p=\nicefrac{1}{3}$.

The \textsc{strict-small} track is tackled by a \textsc{small}-size language model -- 12 encoder layers and 4 decoder layers with hidden size of 384 and with 6 attention heads. The subword vocabulary is reduced to $2^{12} = 4\,096$ items.\footnote{This choice is selected according to \newcite{gowda-may-2020-finding} who recommend to \textit{`\dots use the largest possible vocabulary such that at least 95\% of classes have 100 or more examples in training.'}}

The full list of hyperparameters and implementation details are provided in \cref{app:implementation-details} and in the released source code.\footnote{\url{https://github.com/ltgoslo/boot-bert}}

\subsection{Evaluation}
\label{sec:evaluation}

We utilize the language modeling benchmark suite from the BabyLM challenge \citep{eval-harness, warstadt-et-al-2023-babylm},\footnote{\url{https://github.com/babylm/evaluation-pipeline}} which relies on three conceptually different evaluation tasks:
\begin{enumerate}
    \item The GLUE and SuperGLUE datasets test the ability of a pretrained model to adapt to various language understanding tasks.
    \item BLiMP and BLiMP supplement tasks test the affinity of a model towards grammatical sentences in a completely zero-shot manner.
    \item MSGS measures how much does a pretrained model prefer linguistic generalizations (over surface ones) during finetuning.
\end{enumerate}

\noindent
We further elaborate on each of these evaluation suites below.

\paragraph{(Super)GLUE benchmark.}

General Language Understanding Evaluation benchmarks \citep[GLUE and SuperGLUE;][]{wang-etal-2018-glue, NEURIPS2019_4496bf24} are arguably the most common ways of evaluating the language-understanding and transfer-learning capabilities of language models. The BabyLM challenge uses a subset of 10 (Super)GLUE tasks, detailed in \cref{app:glue}. We employ the standard way of finetuning masked language models on these datasets, as introduced in BERT \citep{devlin-etal-2019-bert}. More details about the finetuning processes are given in \cref{app:implementation-details}.

As we use the BabyLM version of GLUE, our results cannot be directly compared with previous literature -- the dataset samples are filtered to not contain out-of-vocabulary words and some of the employed metrics differ from the original recommendations \citep{wang-etal-2018-glue, NEURIPS2019_4496bf24}. We opted to adhere to the BabyLM version to be compatible with other works in this challenge. However, in order to reliably compare our models, we decided to depart from BabyLM and to divide the training set in 90:10 ratio into a new training and development split; the former validation set is then used as a held-out split.\footnote{The BabyLM pipeline unfortunately uses identical validation and test sets, which might yield overly optimistic results due to overfitting during hyperparameter optimization.} 

\paragraph{BLiMP.}

When using any finetuning approach, it is unclear how to disentangle the innate language understanding from the knowledge learned during the second-stage supervised finetuning \cite{10.1162/coli_a_00422}. In contrast, the Benchmark of Linguistic Minimal Pairs \citep[BLiMP;][]{warstadt-etal-2020-blimp-benchmark} attempts to measure the linguistic knowledge of a language model in a zero-shot manner -- without any additional training. Each pair of sentences in BLiMP differs minimally on the surface level, but only one of the sentences is grammatically valid. We can use the intrinsic ability of language models to assign a probability to every sentence and test how often a language model assigns a higher probability to the correct sentence \citep{wang-cho-2019-bert, salazar-etal-2020-masked}.

As detailed in \cref{app:blimp}, the results on BLiMP greatly depend on temperature scaling \citep{10.5555/3305381.3305518}. Thus, to fairly compare different types of language models, we employ an alternative approach to evaluating BLiMP: we report the accuracies that are achieved with the optimal temperature for every language model; the reasoning is explained in \cref{app:blimp}.

The BabyLM challenge also comes with an additional `BLiMP supplement' held-out set with five additional diagnostic tasks. To comply with the held-out spirit of these tasks, we keep the temperature values calibrated for BLiMP, even though this results in suboptimal performance (\cref{app:blimp}).

\paragraph{MSGS.} The diagnostic dataset called Mixed Signals Generalization Set \citep[MSGS;][]{warstadt-etal-2020-learning} measures whether a pretrained model prefers linguistic or surface generalizations. The experiments follow \textit{the poverty of the stimulus design} \citep{pmid21702842} -- to first finetune a model on ambiguous data (consistent with both linguistic and surface explanations) and then test it on non-ambiguous data to see if it prefers the linguistic generalization. 

We use the filtered MSGS datasets with no \textit{inoculation} in the training set, as provided by the BabyLM challenge. Similarly to (Super)GLUE, we avoid the BabyLM approach that validates and tests on the same split -- instead, to obtain a reliable comparison, we roughly follow the original work \citep{warstadt-etal-2020-learning} and use three learning rates: ($1\cdot10^{-5}$, $2\cdot10^{-5}$, and $3\cdot10^{-5}$), five random seeds, batch size of 16 and finetune for 5 epochs without early-stopping; then we report the mean and standard deviation statistics on the 6 non-ambiguous and non-control test datasets, measuring the Matthew's correlation coefficient \citep[which is renamed to the Linguistic Bias Score (LBS) in MSGS]{MATTHEWS1975442}.

\subsection{Results}

The overall averaged results for all four evaluation suits are given in \cref{tab:metrics}. Apart from evaluating masked autoencoders trained with latent bootstrapping (BootBERTs), as described in \cref{sec:method}, we evaluate the three baseline language models provided by the organizers of BabyLM challenge: decoder-only OPT \citep{zhang2022opt}, encoder-decoder T5 \citep{2020t5} and encoder-only RoBERTa language models \citep{liu2019roberta}. As we base our models on the LTG-BERT architecture \citep{samuel-etal-2023-trained}, we follow recommendations of the authors and also pretrain LTG-BERTs to get a strong and comparable baseline.

In addition to the averaged results, we also provide fine-grained (Super)GLUE scores in \cref{tab:glue} and a visualization of the full distribution of MSGS scores in \cref{fig:msgs} and in \cref{app:msgs} (given the high variation of the aggregated MSGS results). The tables contain the mean and standard deviation statistics over 5 (respectively 15) runs. More details about the BLiMP and BLiMP supplement scores are given in \cref{app:blimp}.

\renewcommand{\arraystretch}{1.4}

\begin{table}
\resizebox{\columnwidth}{!}{%
\begin{tabular}{@{}l@{\hspace{3em}}rrrr@{}}
\toprule
\textbf{Model}           & \textbf{GLUE} & \textbf{MSGS}        & \textbf{BLiMP}       & \textbf{Supplement}  \\ \midrule
\multicolumn{5}{@{}l}{\footnotesize{\textsc{strict} (100M words)}}                                                                              \\
\hspace{1em}OPT\textsubscript{125m}     & 73.0$^{\pm3.9}$  & -44.4$^{\pm8.5}$ &       77.8               &                 67.5     \\
\hspace{1em}RoBERTa\textsubscript{base} &  74.3$^{\pm0.6}$ & -66.4$^{\pm26.6}$  &          76.2            &               63.8       \\
\hspace{1em}T5\textsubscript{base}      &         75.3$^{\pm1.1}$            &               -56.5$^{\pm6.7}$       &           83.6            &          71.8            \\
\hspace{1em}LTG-BERT\textsubscript{base}     & 77.8$^{\pm1.4}$  & \textbf{-43.2}$^{\pm11.0}$ &         \textbf{87.2}            &        \textbf{77.6}                \\ \midrule
\hspace{1em}BootBERT\textsubscript{base}    &  \textbf{79.2}$^{\pm1.5}$ & -67.9$^{\pm12.6}$ &           86.3            &           72.2         \\ [1em]
\multicolumn{5}{@{}l}{\footnotesize{\textsc{strict-small} (10M words)}}                                                                         \\
\hspace{1em}OPT\textsubscript{125m}     & 68.3$^{\pm3.3}$ & -63.8$^{\pm9.6}$ & 69.2 & 60.2 \\
\hspace{1em}RoBERTa\textsubscript{base} & 72.2$^{\pm1.9}$ & -66.7$^{\pm11.9}$ & 68.1 & 60.5 \\
\hspace{1em}T5-base      & 64.7$^{\pm1.3}$ & -68.4$^{\pm7.1}$ & 59.9 & 48.6 \\
\hspace{1em}LTG-BERT\textsubscript{small}     & 74.5$^{\pm1.5}$ & \textbf{-42.6}$^{\pm34.8}$ & 80.9 & \textbf{70.3} \\ \midrule
\hspace{1em}BootBERT\textsubscript{small}     & \textbf{74.9}$^{\pm3.4}$ & -76.6$^{\pm10.2}$ & \textbf{82.2} & 65.6 \\ \bottomrule
\end{tabular}%
}
\caption{The overall average scores for the four evaluation suites: (Super)GLUE, MSGS, BLiMP and BLiMP supplement. The (Super)GLUE and MSGS columns show the mean and standard deviation statistics across multiple runs. The best results for each track are typeset in bold. For a more complete view, the full distribution of the MSGS results is plotted in \cref{fig:msgs}.}
\label{tab:metrics}
\end{table}

\begin{table*}
\resizebox{\textwidth}{!}{%
\begin{tabular}{@{}lrrrrrrrrrrrr@{}}
\toprule
\textbf{Model} & \textbf{BoolQ} & \textbf{CoLA} & \textbf{MNLI\textsubscript{m}} & \textbf{MNLI\textsubscript{mm}} & \textbf{MRPC} & \textbf{MultiRC} & \textbf{QNLI} & \textbf{QQP} & \textbf{RTE} & \textbf{SST-2} & \textbf{WSC} & \textbf{All} \\ \midrule
\multicolumn{5}{@{}l}{\footnotesize{\textsc{strict} (100M words)}}    \\
\hspace{1em}OPT\textsubscript{125m}       & 66.4$^{\pm0.7}$ & 74.9$^{\pm0.6}$ & 75.7$^{\pm0.3}$ & 77.0$^{\pm0.3}$ & 81.9$^{\pm0.7}$ & 61.5$^{\pm0.8}$ & 82.8$^{\pm0.8}$ & 84.3$^{\pm0.1}$ & 58.6$^{\pm2.9}$ & 87.7$^{\pm0.7}$ & 52.3$^{\pm12.5}$ & 73.0$^{\pm3.9}$ \\
\hspace{1em}RoBERTa\textsubscript{base}   & 67.7$^{\pm0.7}$ & 75.6$^{\pm0.3}$ & 77.4$^{\pm0.4}$ & 78.3$^{\pm0.3}$ & 84.0$^{\pm0.5}$ & 64.3$^{\pm0.5}$ & 83.6$^{\pm0.2}$ & 85.5$^{\pm0.2}$ & 50.7$^{\pm1.5}$ & 88.3$^{\pm0.6}$ & \textbf{61.4}$^{\pm0.0}$ & 74.3$^{\pm0.6}$ \\
\hspace{1em}T5\textsubscript{base}        & 67.7$^{\pm1.5}$ & 76.7$^{\pm0.9}$ & 77.9$^{\pm0.3}$ & 78.7$^{\pm0.3}$ & 85.2$^{\pm1.1}$ & 65.7$^{\pm0.8}$ & 84.7$^{\pm0.9}$ & 86.2$^{\pm0.1}$ & 55.4$^{\pm2.2}$ & 89.0$^{\pm0.8}$ & 61.0$^{\pm1.1}$ & 75.3$^{\pm1.1}$ \\
\hspace{1em}LTG-BERT\textsubscript{base}      &  68.1$^{\pm0.4}$ & \textbf{82.8}$^{\pm0.4}$ & 83.4$^{\pm0.3}$ & 83.1$^{\pm0.2}$ & 84.3$^{\pm0.7}$ & \textbf{71.2}$^{\pm0.9}$ & 89.3$^{\pm0.3}$ & 87.9$^{\pm0.2}$ & 55.2$^{\pm2.7}$ & \textbf{91.9}$^{\pm0.6}$ & 58.6$^{\pm3.5}$ & 77.8$^{\pm1.4}$ \\ \midrule
\hspace{1em}BootBERT\textsubscript{base}          &  \textbf{72.4}$^{\pm1.2}$ & 81.6$^{\pm0.6}$ & \textbf{84.7}$^{\pm0.3}$ & \textbf{84.7}$^{\pm0.3}$ & \textbf{89.1}$^{\pm0.3}$ & 70.7$^{\pm1.2}$ & \textbf{91.2}$^{\pm0.4}$ & \textbf{88.1}$^{\pm0.1}$ & \textbf{57.2}$^{\pm3.5}$ & 91.8$^{\pm0.8}$ & 60.2$^{\pm2.7}$ & \textbf{79.2}$^{\pm1.5}$ \\ [1em]

\multicolumn{5}{@{}l}{\footnotesize{\textsc{strict-small} (10M words)}}    \\
\hspace{1em}OPT\textsubscript{125m} & 66.2$^{\pm1.5}$ & 69.0$^{\pm0.5}$ & 69.5$^{\pm0.2}$ & 71.0$^{\pm0.5}$ & 80.0$^{\pm1.8}$ & 56.5$^{\pm2.0}$ & 71.5$^{\pm0.7}$ & 80.3$^{\pm0.3}$ & 51.3$^{\pm2.1}$ & 85.4$^{\pm0.9}$ & 50.8$^{\pm10.3}$ & 68.3$^{\pm3.3}$ \\

\hspace{1em}RoBERTa\textsubscript{base} & 65.8$^{\pm2.9}$ & 70.4$^{\pm0.4}$ & 72.5$^{\pm0.4}$ & 74.4$^{\pm0.3}$ & 82.2$^{\pm0.4}$ & 61.2$^{\pm1.5}$ & 80.3$^{\pm0.7}$ & 83.5$^{\pm0.2}$ & \textbf{56.8}$^{\pm5.5}$ & 85.6$^{\pm0.3}$ & \textbf{61.7}$^{\pm0.5}$ & 72.2$^{\pm1.9}$ \\

\hspace{1em}T5\textsubscript{base} & 63.4$^{\pm1.6}$ & 69.4$^{\pm0.1}$ & 57.3$^{\pm0.8}$ & 58.6$^{\pm1.1}$ & 81.4$^{\pm0.6}$ & 48.4$^{\pm1.4}$ & 64.3$^{\pm0.9}$ & 76.8$^{\pm0.3}$ & 52.7$^{\pm2.4}$ & 79.4$^{\pm1.0}$ & 60.0$^{\pm2.2}$ & 64.7$^{\pm1.3}$ \\

\hspace{1em}LTG-BERT\textsubscript{small} & 64.8$^{\pm2.1}$ & \textbf{77.6}$^{\pm0.8}$ & 78.0$^{\pm0.2}$ & 78.8$^{\pm0.4}$ & 82.3$^{\pm0.4}$ & 64.1$^{\pm0.3}$ & 85.0$^{\pm0.2}$ & 85.8$^{\pm0.2}$ & 53.7$^{\pm4.1}$ & \textbf{88.8}$^{\pm0.8}$ &  60.5$^{\pm1.0}$ & 74.5$^{\pm1.5}$ \\ \midrule

\hspace{1em}BootBERT\textsubscript{small} & \textbf{67.6}$^{\pm2.7}$ & 75.3$^{\pm1.4}$ & \textbf{79.2}$^{\pm0.3}$ & \textbf{80.0}$^{\pm0.2}$ & \textbf{83.2}$^{\pm1.5}$ & \textbf{65.2}$^{\pm0.9}$ & \textbf{86.2}$^{\pm0.4}$ & \textbf{86.6}$^{\pm0.1}$ & 54.7$^{\pm5.3}$ & 88.2$^{\pm0.8}$ & 57.3$^{\pm9.2}$ & \textbf{74.9}$^{\pm3.4}$ \\
\bottomrule
\end{tabular}%
}
\caption{The BabyLM-flavored (Super)GLUE results of language models in the \textsc{strict} track and \textsc{strict-small} track. We present the mean and standard deviation statistics over 5 finetuning runs (initialized with different random seeds) and boldface the best mean-result.}
\label{tab:glue}
\end{table*}

\section{Discussion}

\paragraph{LTG-BERT performance.} Our results confirm the findings by \newcite{samuel-etal-2023-trained} who introduced the improved language modeling architecture called LTG-BERT. These models perform drastically better than the OPT, RoBERTa and T5 baselines pretrained on the same low-resource BabyLM corpus; the performance is improved across all evaluation suites -- GLUE, MSGS, BLiMP as well as the BLiMP supplemental data -- and across both \textsc{strict} and \textsc{strict-small} tracks. LTG-BERT has also been used as the backbone of recent Norwegian language models trained on large amounts of data \citep{samuel-etal-2023-norbench}, which demonstrates that LM methods developed for efficient training are also beneficial for large-scale training.

\paragraph{Self-supervised learning.} When we compare BootBERT to the LTG-BERT baseline, we can see that the latent bootstrapping approach leads to a substantially better performance when finetuned on (Super)GLUE in the \textsc{strict} track and to a slightly better performance in the \textsc{strict-small} track. Specifically on the biggest and arguably most robust GLUE task, MNLI, the accuracy is better by 1.3/1.6 percentage points in the \textsc{strict} track and by 1.2/1.2pp in the \textsc{strict-small} track. The overall average (Super)GLUE score is better by 1.4pp and by 0.4pp, respectively. This shows that language models pretrained with this approach are a good option for downstream tasks.

\begin{figure}[!t]
    \includegraphics[width=\columnwidth]{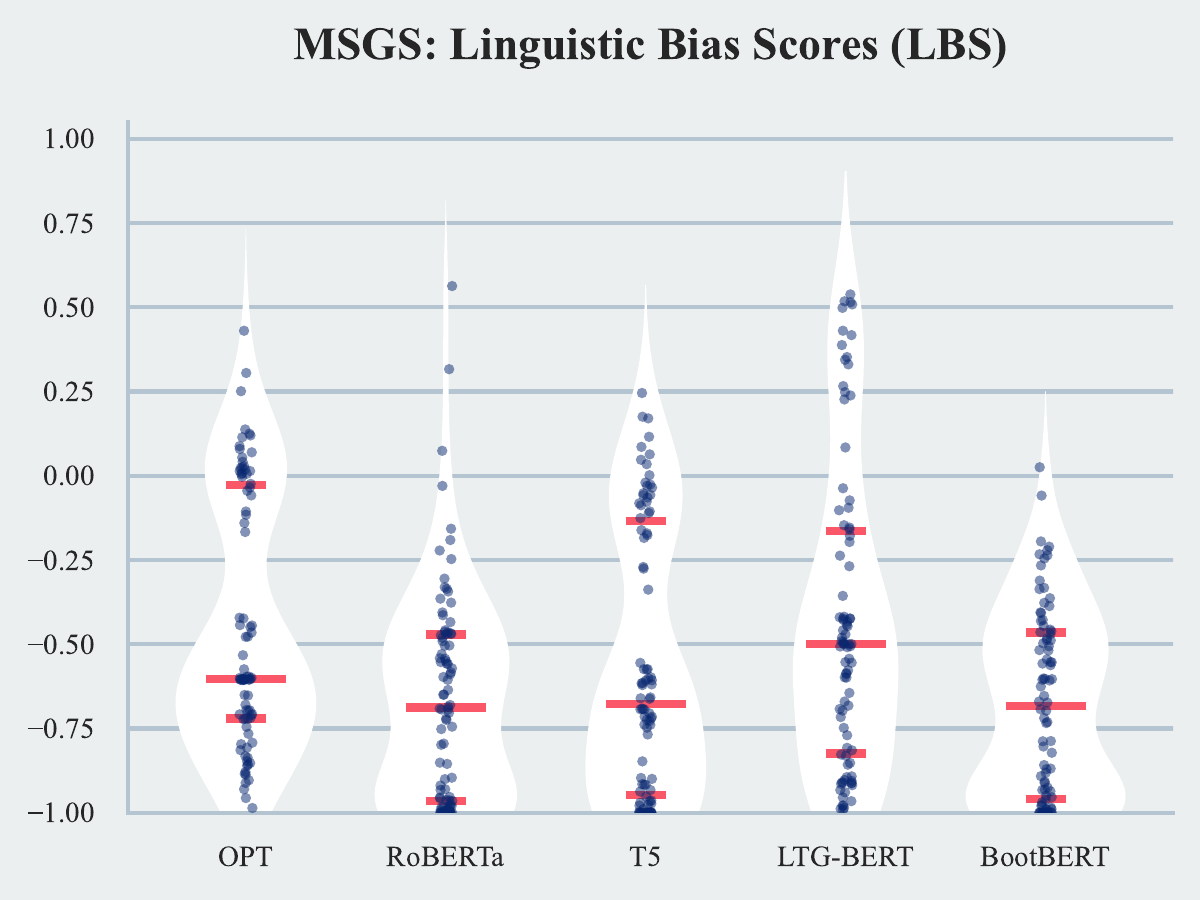}
    \caption{The Linguistic Bias Scores (LBS) of language models pretrained on the \textsc{strict} dataset. These plots show the distribution of the LBS scores across 15 evaluation runs (3 learning rates $\times$ 5 random seeds) for each of the 6 non-ambiguous test datasets (90 values in total for each model). The red horizontal lines highlight the first, second (median) and third quartile. The overall negative scores show that none of the tested models prefers linguistic features over the surface ones.}
    \label{fig:msgs}
\end{figure}

The ability of linguistic generalization, as measured by the linguistic bias scores in MSGS, is substantially worse in BootBERT than in the LTG-BERT baseline, as evident from \cref{fig:msgs}. A more detailed analysis in \cref{app:msgs} reveals that this holds for both BabyLM tracks -- but the difference is mainly due to the fact that LTG-BERT reliably prefers the linguistic feature `is the main verb in “ing” form?', other tests are relatively similar for both types of models. It is unclear what part of latent bootstrapping causes this difference. 

The results on the BLiMP-based benchmarks are mixed but overall worse when comparing BootBERT with the LTG-BERT baseline. This is possibly because of the utilization of two conflicting training objectives in BootBERT -- intuitively, pure language-modeling-based training should have an advantage on benchmarks that rely on sentence likelihood.

In conclusion, these low-resource experiments suggest that \textbf{the advantage of latent bootstrapping for natural language is not as great as the advantage that has been previously demonstrated for computer vision}. We believe that this is because the atomic units of text, subword tokens, can provide much more semantically rich signal when compared to the atomic units of images, pixels. Thus there is not a large need for bootstrapping a rich signal from a teacher; instead, the standard language modeling comes with a training objective that is simple and provides enough signal, while suffering from issues like representation collapse.

\paragraph{The shared task results.} The official DynaBench results for BabyLM can be found in \cref{app:dynabench}. Our system ranks high when evaluated on GLUE (first and second place) and on BLiMP (second and first place) in the \textsc{strict} and \textsc{strict-small} tracks, respectively. As discussed earlier, BootBERT strongly prefers the surface features over the linguistic features and thus places low on the MSGS benchmark (third and last place), which also hurts the overall ranking of our system (third an seventh). 
Note however that this evaluation is not using a proper train/development/test split and it does not account for high variation of some metrics (MSGS in particular), which is why we have used an alternative evaluation in the rest of this paper.

\paragraph{Computational cost of latent bootstrapping.} It is important to note that latent bootstrapping comes with an increased computational cost because of an additional forward pass through the mean teacher; which roughly equates to a 50\% increase in pretraining time. Thus, it should be carefully considered whether the potential benefits of bootstrapping are worth this cost. That being said, this method does not bear any additional cost during finetuning nor inference, which might justify it in some cases.

\section{Related work}

\paragraph{Self-supervised learning.} Our work is greatly inspired by the `bootstrap your own latent' approach \citep[BYOL;][]{10.5555/3495724.3497510}, which introduced the bootstrapping feedback loop between a student and a mean teacher network. BYOL by itself can be considered an example of contrastive learning \citep{hjelm2018learning, oord2019representation, 10.5555/3524938.3525087, 9157636} without negative instances. Another important aspect of BYOL is the usage of a `mean teacher', a slow-moving average of a student network, which is a term coined by \newcite{10.5555/3294771.3294885}.

Many methods of visual representation learning adopted the bootstrapping approach and further improved its parts \citep{Chen_2021_CVPR, pmlr-v139-zbontar21a, bardes2022vicreg, He_2022_CVPR}. In particular, our work bears similarities with the recently introduces `image-based joint-embedding predictive architecture' \citep[I-JEPA;][]{assran2023self}, which also trains a masked autoencoder student network to predict the contextualized embeddings of an unmasked mean teacher. While mostly used for the image domain, \textit{data2vec} method showed that latent bootstrapping can also be successfully applied to text \citep{pmlr-v162-baevski22a}.

\paragraph{Efficient language modeling.} The necessity of pretraining modern language models on large corpora were questioned in CamemBERT \citep{martin-etal-2020-camembert} and the effect of corpus size has been then thoroughly studied in \newcite{micheli-etal-2020-importance}, \newcite{zhang-etal-2021-need} as well as in \newcite{https://doi.org/10.48550/arxiv.2203.15556}. \newcite{samuel-etal-2023-trained} introduced the LTG-BERT -- an improved language model optimized for pretraining on a low-resource corpus. They showed that a well-tuned language model can match the performance of BERT even when it is pretrained only on a small 100-million-word British National Corpus (BNC). We base our approach on this model due to the apparent similarity of the BabyLM training corpus to BNC.

\section{Conclusion}

In this paper, we presented a masked autoencoder language model trained with latent bootstrapping, an alternative self-supervised learning method. We showed that when pretrained on a low-resource corpus, the results of this method are varied -- compared to a masked language modeling baseline, the performance is clearly better on (Super)GLUE, but worse on MSGS and mixed on BLiMP. We believe that it makes a promising alternative to traditional language modeling methods, but its reliable and effective utilization requires future work.

\section*{Acknowledgements}
This paper would not be possible without the endless support and incredibly useful feedback from Andrey Kutuzov, Erik Velldal and Lilja Øvrelid from the Language Technology Group at the University of Oslo.

The efforts described in the current paper were funded by the HPLT project (High Performance Language Technologies; coordinated by Charles University). The computations were performed on resources provided through Sigma2 -- the national research infrastructure provider for High-Performance Computing and large-scale data storage in Norway.

\bibliography{anthology,custom}

\begin{thebibliography}{68}
\expandafter\ifx\csname natexlab\endcsname\relax\def\natexlab#1{#1}\fi

\bibitem[{Abdelali et~al.(2014)Abdelali, Guzman, Sajjad, and
  Vogel}]{abdelali-etal-2014-qed}
Ahmed Abdelali, Francisco Guzman, Hassan Sajjad, and Stephan Vogel. 2014.
\newblock The {AMARA} corpus: Building parallel language resources for the
  educational domain.
\newblock In \emph{Proceedings of the Ninth International Conference on
  Language Resources and Evaluation (LREC'14)}, Reykjavik, Iceland. European
  Language Resources Association (ELRA).

\bibitem[{Assran et~al.(2023)Assran, Duval, Misra, Bojanowski, Vincent, Rabbat,
  LeCun, and Ballas}]{assran2023self}
Mahmoud Assran, Quentin Duval, Ishan Misra, Piotr Bojanowski, Pascal Vincent,
  Michael Rabbat, Yann LeCun, and Nicolas Ballas. 2023.
\newblock Self-supervised learning from images with a joint-embedding
  predictive architecture.
\newblock \emph{arXiv preprint arXiv:2301.08243}.

\bibitem[{Baevski et~al.(2022)Baevski, Hsu, Xu, Babu, Gu, and
  Auli}]{pmlr-v162-baevski22a}
Alexei Baevski, Wei-Ning Hsu, Qiantong Xu, Arun Babu, Jiatao Gu, and Michael
  Auli. 2022.
\newblock \href {https://proceedings.mlr.press/v162/baevski22a.html} {data2vec:
  A general framework for self-supervised learning in speech, vision and
  language}.
\newblock In \emph{Proceedings of the 39th International Conference on Machine
  Learning}, volume 162 of \emph{Proceedings of Machine Learning Research},
  pages 1298--1312. PMLR.

\bibitem[{Bar-Haim et~al.(2006)Bar-Haim, Dagan, Dolan, Ferro, and
  Giampiccolo}]{rte2}
Roy Bar-Haim, Ido Dagan, Bill Dolan, Lisa Ferro, and Danilo Giampiccolo. 2006.
\newblock The second pascal recognising textual entailment challenge.
\newblock \emph{Proceedings of the Second PASCAL Challenges Workshop on
  Recognising Textual Entailment}.

\bibitem[{Bardes et~al.(2022)Bardes, Ponce, and LeCun}]{bardes2022vicreg}
Adrien Bardes, Jean Ponce, and Yann LeCun. 2022.
\newblock \href {https://openreview.net/forum?id=xm6YD62D1Ub} {{VICR}eg:
  Variance-invariance-covariance regularization for self-supervised learning}.
\newblock In \emph{International Conference on Learning Representations}.

\bibitem[{Belinkov(2022)}]{10.1162/coli_a_00422}
Yonatan Belinkov. 2022.
\newblock \href {https://doi.org/10.1162/coli_a_00422} {{Probing Classifiers:
  Promises, Shortcomings, and Advances}}.
\newblock \emph{Computational Linguistics}, 48(1):207--219.

\bibitem[{Bentivogli et~al.(2009)Bentivogli, Dagan, Dang, Giampiccolo, and
  Magnini}]{Bentivogli09thefifth}
Luisa Bentivogli, Ido Dagan, Hoa~Trang Dang, Danilo Giampiccolo, and Bernardo
  Magnini. 2009.
\newblock The fifth pascal recognizing textual entailment challenge.
\newblock In \emph{In Proc Text Analysis Conference (TAC’09}.

\bibitem[{Chen et~al.(2020)Chen, Kornblith, Norouzi, and
  Hinton}]{10.5555/3524938.3525087}
Ting Chen, Simon Kornblith, Mohammad Norouzi, and Geoffrey Hinton. 2020.
\newblock A simple framework for contrastive learning of visual
  representations.
\newblock In \emph{Proceedings of the 37th International Conference on Machine
  Learning}, ICML'20. JMLR.org.

\bibitem[{Chen and He(2020)}]{chen2020simsiam}
Xinlei Chen and Kaiming He. 2020.
\newblock Exploring simple siamese representation learning.
\newblock \emph{arXiv preprint arXiv:2011.10566}.

\bibitem[{Chen and He(2021)}]{Chen_2021_CVPR}
Xinlei Chen and Kaiming He. 2021.
\newblock Exploring simple siamese representation learning.
\newblock In \emph{Proceedings of the IEEE/CVF Conference on Computer Vision
  and Pattern Recognition (CVPR)}, pages 15750--15758.

\bibitem[{Clark et~al.(2019)Clark, Lee, Chang, Kwiatkowski, Collins, and
  Toutanova}]{clark-etal-2019-boolq}
Christopher Clark, Kenton Lee, Ming-Wei Chang, Tom Kwiatkowski, Michael
  Collins, and Kristina Toutanova. 2019.
\newblock \href {https://doi.org/10.18653/v1/N19-1300} {{B}ool{Q}: Exploring
  the surprising difficulty of natural yes/no questions}.
\newblock In \emph{Proceedings of the 2019 Conference of the North {A}merican
  Chapter of the Association for Computational Linguistics: Human Language
  Technologies, Volume 1 (Long and Short Papers)}, pages 2924--2936,
  Minneapolis, Minnesota. Association for Computational Linguistics.

\bibitem[{Dagan et~al.(2006)Dagan, Glickman, and Magnini}]{10.1007/11736790_9}
Ido Dagan, Oren Glickman, and Bernardo Magnini. 2006.
\newblock The pascal recognising textual entailment challenge.
\newblock In \emph{Machine Learning Challenges. Evaluating Predictive
  Uncertainty, Visual Object Classification, and Recognising Tectual
  Entailment}, pages 177--190, Berlin, Heidelberg. Springer Berlin Heidelberg.

\bibitem[{Devlin et~al.(2019)Devlin, Chang, Lee, and
  Toutanova}]{devlin-etal-2019-bert}
Jacob Devlin, Ming-Wei Chang, Kenton Lee, and Kristina Toutanova. 2019.
\newblock \href {https://doi.org/10.18653/v1/N19-1423} {{BERT}: Pre-training of
  deep bidirectional transformers for language understanding}.
\newblock In \emph{Proceedings of the 2019 Conference of the North {A}merican
  Chapter of the Association for Computational Linguistics: Human Language
  Technologies, Volume 1 (Long and Short Papers)}, pages 4171--4186,
  Minneapolis, Minnesota. Association for Computational Linguistics.

\bibitem[{Dolan and Brockett(2005)}]{dolan-brockett-2005-automatically}
William~B. Dolan and Chris Brockett. 2005.
\newblock \href {https://aclanthology.org/I05-5002} {Automatically constructing
  a corpus of sentential paraphrases}.
\newblock In \emph{Proceedings of the Third International Workshop on
  Paraphrasing ({IWP}2005)}.

\bibitem[{Ettinger(2020)}]{ettinger-2020-bert}
Allyson Ettinger. 2020.
\newblock \href {https://doi.org/10.1162/tacl_a_00298} {What {BERT} is not:
  Lessons from a new suite of psycholinguistic diagnostics for language
  models}.
\newblock \emph{Transactions of the Association for Computational Linguistics},
  8:34--48.

\bibitem[{Gao et~al.(2021)Gao, Tow, Biderman, Black, DiPofi, Foster, Golding,
  Hsu, McDonell, Muennighoff, Phang, Reynolds, Tang, Thite, Wang, Wang, and
  Zou}]{eval-harness}
Leo Gao, Jonathan Tow, Stella Biderman, Sid Black, Anthony DiPofi, Charles
  Foster, Laurence Golding, Jeffrey Hsu, Kyle McDonell, Niklas Muennighoff,
  Jason Phang, Laria Reynolds, Eric Tang, Anish Thite, Ben Wang, Kevin Wang,
  and Andy Zou. 2021.
\newblock \href {https://doi.org/10.5281/zenodo.5371628} {A framework for
  few-shot language model evaluation}.

\bibitem[{Gerlach and Font-Clos(2018)}]{gerlach-2018-gutenberg}
Martin Gerlach and Francesc Font-Clos. 2018.
\newblock \href {https://doi.org/10.48550/ARXIV.1812.08092} {A standardized
  {Project Gutenberg} corpus for statistical analysis of natural language and
  quantitative linguistics}.
\newblock \emph{Computing Research Repository}, arXiv:1812.08092.

\bibitem[{Giampiccolo et~al.(2007)Giampiccolo, Magnini, Dagan, and
  Dolan}]{giampiccolo-etal-2007-third}
Danilo Giampiccolo, Bernardo Magnini, Ido Dagan, and Bill Dolan. 2007.
\newblock \href {https://aclanthology.org/W07-1401} {The third {PASCAL}
  recognizing textual entailment challenge}.
\newblock In \emph{Proceedings of the {ACL}-{PASCAL} Workshop on Textual
  Entailment and Paraphrasing}, pages 1--9, Prague. Association for
  Computational Linguistics.

\bibitem[{Girshick(2015)}]{10.1109/ICCV.2015.169}
Ross Girshick. 2015.
\newblock \href {https://doi.org/10.1109/ICCV.2015.169} {Fast r-cnn}.
\newblock In \emph{Proceedings of the 2015 IEEE International Conference on
  Computer Vision (ICCV)}, ICCV '15, page 1440–1448, USA. IEEE Computer
  Society.

\bibitem[{Gowda and May(2020)}]{gowda-may-2020-finding}
Thamme Gowda and Jonathan May. 2020.
\newblock \href {https://doi.org/10.18653/v1/2020.findings-emnlp.352} {Finding
  the optimal vocabulary size for neural machine translation}.
\newblock In \emph{Findings of the Association for Computational Linguistics:
  EMNLP 2020}, pages 3955--3964, Online. Association for Computational
  Linguistics.

\bibitem[{Grill et~al.(2020)Grill, Strub, Altch\'{e}, Tallec, Richemond,
  Buchatskaya, Doersch, Pires, Guo, Azar, Piot, Kavukcuoglu, Munos, and
  Valko}]{10.5555/3495724.3497510}
Jean-Bastien Grill, Florian Strub, Florent Altch\'{e}, Corentin Tallec,
  Pierre~H. Richemond, Elena Buchatskaya, Carl Doersch, Bernardo~Avila Pires,
  Zhaohan~Daniel Guo, Mohammad~Gheshlaghi Azar, Bilal Piot, Koray Kavukcuoglu,
  R\'{e}mi Munos, and Michal Valko. 2020.
\newblock Bootstrap your own latent a new approach to self-supervised learning.
\newblock In \emph{Proceedings of the 34th International Conference on Neural
  Information Processing Systems}, NIPS'20, Red Hook, NY, USA. Curran
  Associates Inc.

\bibitem[{Guo et~al.(2017{\natexlab{a}})Guo, Pleiss, Sun, and
  Weinberger}]{10.5555/3305381.3305518}
Chuan Guo, Geoff Pleiss, Yu~Sun, and Kilian~Q. Weinberger. 2017{\natexlab{a}}.
\newblock On calibration of modern neural networks.
\newblock In \emph{Proceedings of the 34th International Conference on Machine
  Learning - Volume 70}, ICML'17, page 1321–1330. JMLR.org.

\bibitem[{Guo et~al.(2017{\natexlab{b}})Guo, Pleiss, Sun, and
  Weinberger}]{guo2017calibration}
Chuan Guo, Geoff Pleiss, Yu~Sun, and Kilian~Q Weinberger. 2017{\natexlab{b}}.
\newblock On calibration of modern neural networks.

\bibitem[{He et~al.(2022)He, Chen, Xie, Li, Doll\'ar, and
  Girshick}]{He_2022_CVPR}
Kaiming He, Xinlei Chen, Saining Xie, Yanghao Li, Piotr Doll\'ar, and Ross
  Girshick. 2022.
\newblock Masked autoencoders are scalable vision learners.
\newblock In \emph{Proceedings of the IEEE/CVF Conference on Computer Vision
  and Pattern Recognition (CVPR)}, pages 16000--16009.

\bibitem[{He et~al.(2020)He, Fan, Wu, Xie, and Girshick}]{9157636}
Kaiming He, Haoqi Fan, Yuxin Wu, Saining Xie, and Ross Girshick. 2020.
\newblock \href {https://doi.org/10.1109/CVPR42600.2020.00975} {Momentum
  contrast for unsupervised visual representation learning}.
\newblock In \emph{2020 IEEE/CVF Conference on Computer Vision and Pattern
  Recognition (CVPR)}, pages 9726--9735.

\bibitem[{He et~al.(2021)He, Liu, Gao, and Chen}]{he2021deberta}
Pengcheng He, Xiaodong Liu, Jianfeng Gao, and Weizhu Chen. 2021.
\newblock \href {https://openreview.net/forum?id=XPZIaotutsD} {Deberta:
  Decoding-enhanced bert with disentangled attention}.
\newblock In \emph{International Conference on Learning Representations}.

\bibitem[{Hill et~al.(2016)Hill, Bordes, Chopra, and Weston}]{hill-2016-cbt}
Felix Hill, Antoine Bordes, Sumit Chopra, and Jason Weston. 2016.
\newblock \href {http://arxiv.org/abs/1511.02301} {The {Goldilocks} principle:
  Reading children's books with explicit memory representations}.

\bibitem[{Hjelm et~al.(2019)Hjelm, Fedorov, Lavoie-Marchildon, Grewal, Bachman,
  Trischler, and Bengio}]{hjelm2018learning}
R~Devon Hjelm, Alex Fedorov, Samuel Lavoie-Marchildon, Karan Grewal, Phil
  Bachman, Adam Trischler, and Yoshua Bengio. 2019.
\newblock \href {https://openreview.net/forum?id=Bklr3j0cKX} {Learning deep
  representations by mutual information estimation and maximization}.
\newblock In \emph{International Conference on Learning Representations}.

\bibitem[{Hoffmann et~al.(2022)Hoffmann, Borgeaud, Mensch, Buchatskaya, Cai,
  Rutherford, Casas, Hendricks, Welbl, Clark, Hennigan, Noland, Millican,
  Driessche, Damoc, Guy, Osindero, Simonyan, Elsen, Rae, Vinyals, and
  Sifre}]{https://doi.org/10.48550/arxiv.2203.15556}
Jordan Hoffmann, Sebastian Borgeaud, Arthur Mensch, Elena Buchatskaya, Trevor
  Cai, Eliza Rutherford, Diego de~Las Casas, Lisa~Anne Hendricks, Johannes
  Welbl, Aidan Clark, Tom Hennigan, Eric Noland, Katie Millican, George van~den
  Driessche, Bogdan Damoc, Aurelia Guy, Simon Osindero, Karen Simonyan, Erich
  Elsen, Jack~W. Rae, Oriol Vinyals, and Laurent Sifre. 2022.
\newblock \href {https://doi.org/10.48550/ARXIV.2203.15556} {Training
  compute-optimal large language models}.

\bibitem[{Joshi et~al.(2020)Joshi, Chen, Liu, Weld, Zettlemoyer, and
  Levy}]{joshi-etal-2020-spanbert}
Mandar Joshi, Danqi Chen, Yinhan Liu, Daniel~S. Weld, Luke Zettlemoyer, and
  Omer Levy. 2020.
\newblock \href {https://doi.org/10.1162/tacl_a_00300} {{S}pan{BERT}: Improving
  pre-training by representing and predicting spans}.
\newblock \emph{Transactions of the Association for Computational Linguistics},
  8:64--77.

\bibitem[{Khashabi et~al.(2018)Khashabi, Chaturvedi, Roth, Upadhyay, and
  Roth}]{khashabi-etal-2018-looking}
Daniel Khashabi, Snigdha Chaturvedi, Michael Roth, Shyam Upadhyay, and Dan
  Roth. 2018.
\newblock \href {https://doi.org/10.18653/v1/N18-1023} {Looking beyond the
  surface: A challenge set for reading comprehension over multiple sentences}.
\newblock In \emph{Proceedings of the 2018 Conference of the North {A}merican
  Chapter of the Association for Computational Linguistics: Human Language
  Technologies, Volume 1 (Long Papers)}, pages 252--262, New Orleans,
  Louisiana. Association for Computational Linguistics.

\bibitem[{LeCun(2022)}]{lecun2022path}
Yann LeCun. 2022.
\newblock A path towards autonomous machine intelligence version 0.9. 2,
  2022-06-27.
\newblock \emph{Open Review}, 62.

\bibitem[{Levesque et~al.(2012)Levesque, Davis, and
  Morgenstern}]{10.5555/3031843.3031909}
Hector~J. Levesque, Ernest Davis, and Leora Morgenstern. 2012.
\newblock The winograd schema challenge.
\newblock In \emph{Proceedings of the Thirteenth International Conference on
  Principles of Knowledge Representation and Reasoning}, KR'12, page 552–561.
  AAAI Press.

\bibitem[{Lison and Tiedemann(2016)}]{lison-tiedemann-2016-opensubtitles2016}
Pierre Lison and J{\"o}rg Tiedemann. 2016.
\newblock \href {https://aclanthology.org/L16-1147} {{O}pen{S}ubtitles2016:
  Extracting large parallel corpora from movie and {TV} subtitles}.
\newblock In \emph{Proceedings of the Tenth International Conference on
  Language Resources and Evaluation ({LREC}'16)}, pages 923--929,
  Portoro{\v{z}}, Slovenia. European Language Resources Association (ELRA).

\bibitem[{Liu et~al.(2019)Liu, Ott, Goyal, Du, Joshi, Chen, Levy, Lewis,
  Zettlemoyer, and Stoyanov}]{liu2019roberta}
Yinhan Liu, Myle Ott, Naman Goyal, Jingfei Du, Mandar Joshi, Danqi Chen, Omer
  Levy, Mike Lewis, Luke Zettlemoyer, and Veselin Stoyanov. 2019.
\newblock \href {http://arxiv.org/abs/1907.11692} {Roberta: A robustly
  optimized bert pretraining approach}.

\bibitem[{MacWhinney(2000)}]{macwhinney2000childes}
Brian MacWhinney. 2000.
\newblock \emph{The CHILDES project: The database}, volume~2.
\newblock Psychology Press.

\bibitem[{Martin et~al.(2020)Martin, Muller, Ortiz~Su{\'a}rez, Dupont, Romary,
  de~la Clergerie, Seddah, and Sagot}]{martin-etal-2020-camembert}
Louis Martin, Benjamin Muller, Pedro~Javier Ortiz~Su{\'a}rez, Yoann Dupont,
  Laurent Romary, {\'E}ric de~la Clergerie, Djam{\'e} Seddah, and Beno{\^\i}t
  Sagot. 2020.
\newblock \href {https://doi.org/10.18653/v1/2020.acl-main.645} {{C}amem{BERT}:
  a tasty {F}rench language model}.
\newblock In \emph{Proceedings of the 58th Annual Meeting of the Association
  for Computational Linguistics}, pages 7203--7219, Online. Association for
  Computational Linguistics.

\bibitem[{Matthews(1975)}]{MATTHEWS1975442}
B.W. Matthews. 1975.
\newblock \href {https://doi.org/https://doi.org/10.1016/0005-2795(75)90109-9}
  {Comparison of the predicted and observed secondary structure of t4 phage
  lysozyme}.
\newblock \emph{Biochimica et Biophysica Acta (BBA) - Protein Structure},
  405(2):442--451.

\bibitem[{Meng et~al.(2023)Meng, Krishnan, Wang, Wang, Mao, Fang,
  Ghazvininejad, Han, and Zettlemoyer}]{meng2023representation}
Yu~Meng, Jitin Krishnan, Sinong Wang, Qifan Wang, Yuning Mao, Han Fang, Marjan
  Ghazvininejad, Jiawei Han, and Luke Zettlemoyer. 2023.
\newblock \href {http://arxiv.org/abs/2302.02060} {Representation deficiency in
  masked language modeling}.

\bibitem[{Micheli et~al.(2020)Micheli, d{'}Hoffschmidt, and
  Fleuret}]{micheli-etal-2020-importance}
Vincent Micheli, Martin d{'}Hoffschmidt, and Fran{\c{c}}ois Fleuret. 2020.
\newblock \href {https://doi.org/10.18653/v1/2020.emnlp-main.632} {On the
  importance of pre-training data volume for compact language models}.
\newblock In \emph{Proceedings of the 2020 Conference on Empirical Methods in
  Natural Language Processing (EMNLP)}, pages 7853--7858, Online. Association
  for Computational Linguistics.

\bibitem[{Radford et~al.(2018)Radford, Narasimhan, Salimans, and
  Sutskever}]{Radford2018ImprovingLU}
Alec Radford, Karthik Narasimhan, Tim Salimans, and Ilya Sutskever. 2018.
\newblock \href {https://api.semanticscholar.org/CorpusID:49313245} {Improving
  language understanding by generative pre-training}.

\bibitem[{Raffel et~al.(2020)Raffel, Shazeer, Roberts, Lee, Narang, Matena,
  Zhou, Li, and Liu}]{2020t5}
Colin Raffel, Noam Shazeer, Adam Roberts, Katherine Lee, Sharan Narang, Michael
  Matena, Yanqi Zhou, Wei Li, and Peter~J. Liu. 2020.
\newblock \href {http://jmlr.org/papers/v21/20-074.html} {Exploring the limits
  of transfer learning with a unified text-to-text transformer}.
\newblock \emph{Journal of Machine Learning Research}, 21(140):1--67.

\bibitem[{Rajpurkar et~al.(2016)Rajpurkar, Zhang, Lopyrev, and
  Liang}]{rajpurkar-etal-2016-squad}
Pranav Rajpurkar, Jian Zhang, Konstantin Lopyrev, and Percy Liang. 2016.
\newblock \href {https://doi.org/10.18653/v1/D16-1264} {{SQ}u{AD}: 100,000+
  questions for machine comprehension of text}.
\newblock In \emph{Proceedings of the 2016 Conference on Empirical Methods in
  Natural Language Processing}, pages 2383--2392, Austin, Texas. Association
  for Computational Linguistics.

\bibitem[{Salazar et~al.(2020)Salazar, Liang, Nguyen, and
  Kirchhoff}]{salazar-etal-2020-masked}
Julian Salazar, Davis Liang, Toan~Q. Nguyen, and Katrin Kirchhoff. 2020.
\newblock \href {https://doi.org/10.18653/v1/2020.acl-main.240} {Masked
  language model scoring}.
\newblock In \emph{Proceedings of the 58th Annual Meeting of the Association
  for Computational Linguistics}, pages 2699--2712, Online. Association for
  Computational Linguistics.

\bibitem[{Samuel et~al.(2023{\natexlab{a}})Samuel, Kutuzov, {\O}vrelid, and
  Velldal}]{samuel-etal-2023-trained}
David Samuel, Andrey Kutuzov, Lilja {\O}vrelid, and Erik Velldal.
  2023{\natexlab{a}}.
\newblock \href {https://aclanthology.org/2023.findings-eacl.146} {Trained on
  100 million words and still in shape: {BERT} meets {B}ritish {N}ational
  {C}orpus}.
\newblock In \emph{Findings of the Association for Computational Linguistics:
  EACL 2023}, pages 1954--1974, Dubrovnik, Croatia. Association for
  Computational Linguistics.

\bibitem[{Samuel et~al.(2023{\natexlab{b}})Samuel, Kutuzov, Touileb, Velldal,
  {\O}vrelid, R{\o}nningstad, Sigdel, and
  Palatkina}]{samuel-etal-2023-norbench}
David Samuel, Andrey Kutuzov, Samia Touileb, Erik Velldal, Lilja {\O}vrelid,
  Egil R{\o}nningstad, Elina Sigdel, and Anna Palatkina. 2023{\natexlab{b}}.
\newblock \href {https://aclanthology.org/2023.nodalida-1.61} {{N}or{B}ench
  {--} a benchmark for {N}orwegian language models}.
\newblock In \emph{Proceedings of the 24th Nordic Conference on Computational
  Linguistics (NoDaLiDa)}, pages 618--633, T{\'o}rshavn, Faroe Islands.
  University of Tartu Library.

\bibitem[{Shazeer(2020)}]{DBLP:journals/corr/abs-2002-05202}
Noam Shazeer. 2020.
\newblock \href {http://arxiv.org/abs/2002.05202} {{GLU} variants improve
  transformer}.
\newblock \emph{CoRR}, abs/2002.05202.

\bibitem[{Shleifer and Ott(2022)}]{shleifer2022normformer}
Sam Shleifer and Myle Ott. 2022.
\newblock \href {https://openreview.net/forum?id=GMYWzWztDx5} {Normformer:
  Improved transformer pretraining with extra normalization}.

\bibitem[{Socher et~al.(2013)Socher, Perelygin, Wu, Chuang, Manning, Ng, and
  Potts}]{socher-etal-2013-recursive}
Richard Socher, Alex Perelygin, Jean Wu, Jason Chuang, Christopher~D. Manning,
  Andrew Ng, and Christopher Potts. 2013.
\newblock \href {https://aclanthology.org/D13-1170} {Recursive deep models for
  semantic compositionality over a sentiment treebank}.
\newblock In \emph{Proceedings of the 2013 Conference on Empirical Methods in
  Natural Language Processing}, pages 1631--1642, Seattle, Washington, USA.
  Association for Computational Linguistics.

\bibitem[{Stolcke et~al.(2000)Stolcke, Ries, Coccaro, Shriberg, Bates,
  Jurafsky, Taylor, Martin, Meteer, and Van Ess-Dykema}]{Stolcke-etal:2000}
Andreas Stolcke, Klaus Ries, Noah Coccaro, Elizabeth Shriberg, Rebecca Bates,
  Daniel Jurafsky, Paul Taylor, Rachel Martin, Marie Meteer, and Carol Van
  Ess-Dykema. 2000.
\newblock Dialogue act modeling for automatic tagging and recognition of
  conversational speech.
\newblock \emph{Computational Linguistics}, 26(3):339--371.

\bibitem[{Tarvainen and Valpola(2017)}]{10.5555/3294771.3294885}
Antti Tarvainen and Harri Valpola. 2017.
\newblock Mean teachers are better role models: Weight-averaged consistency
  targets improve semi-supervised deep learning results.
\newblock In \emph{Proceedings of the 31st International Conference on Neural
  Information Processing Systems}, NIPS'17, page 1195–1204, Red Hook, NY,
  USA. Curran Associates Inc.

\bibitem[{Tenney et~al.(2019)Tenney, Xia, Chen, Wang, Poliak, McCoy, Kim,
  Durme, Bowman, Das, and Pavlick}]{tenney2018what}
Ian Tenney, Patrick Xia, Berlin Chen, Alex Wang, Adam Poliak, R~Thomas McCoy,
  Najoung Kim, Benjamin~Van Durme, Sam Bowman, Dipanjan Das, and Ellie Pavlick.
  2019.
\newblock \href {https://openreview.net/forum?id=SJzSgnRcKX} {What do you learn
  from context? probing for sentence structure in contextualized word
  representations}.
\newblock In \emph{International Conference on Learning Representations}.

\bibitem[{van~den Oord et~al.(2019)van~den Oord, Li, and
  Vinyals}]{oord2019representation}
Aaron van~den Oord, Yazhe Li, and Oriol Vinyals. 2019.
\newblock \href {http://arxiv.org/abs/1807.03748} {Representation learning with
  contrastive predictive coding}.

\bibitem[{Vaswani et~al.(2017)Vaswani, Shazeer, Parmar, Uszkoreit, Jones,
  Gomez, Kaiser, and Polosukhin}]{NIPS2017_3f5ee243}
Ashish Vaswani, Noam Shazeer, Niki Parmar, Jakob Uszkoreit, Llion Jones,
  Aidan~N Gomez, \L~ukasz Kaiser, and Illia Polosukhin. 2017.
\newblock \href
  {https://proceedings.neurips.cc/paper_files/paper/2017/file/3f5ee243547dee91fbd053c1c4a845aa-Paper.pdf}
  {Attention is all you need}.
\newblock In \emph{Advances in Neural Information Processing Systems},
  volume~30. Curran Associates, Inc.

\bibitem[{Wang and Cho(2019)}]{wang-cho-2019-bert}
Alex Wang and Kyunghyun Cho. 2019.
\newblock \href {https://doi.org/10.18653/v1/W19-2304} {{BERT} has a mouth, and
  it must speak: {BERT} as a {M}arkov random field language model}.
\newblock In \emph{Proceedings of the Workshop on Methods for Optimizing and
  Evaluating Neural Language Generation}, pages 30--36, Minneapolis, Minnesota.
  Association for Computational Linguistics.

\bibitem[{Wang et~al.(2019)Wang, Pruksachatkun, Nangia, Singh, Michael, Hill,
  Levy, and Bowman}]{NEURIPS2019_4496bf24}
Alex Wang, Yada Pruksachatkun, Nikita Nangia, Amanpreet Singh, Julian Michael,
  Felix Hill, Omer Levy, and Samuel Bowman. 2019.
\newblock \href
  {https://proceedings.neurips.cc/paper/2019/file/4496bf24afe7fab6f046bf4923da8de6-Paper.pdf}
  {Superglue: A stickier benchmark for general-purpose language understanding
  systems}.
\newblock In \emph{Advances in Neural Information Processing Systems},
  volume~32. Curran Associates, Inc.

\bibitem[{Wang et~al.(2018)Wang, Singh, Michael, Hill, Levy, and
  Bowman}]{wang-etal-2018-glue}
Alex Wang, Amanpreet Singh, Julian Michael, Felix Hill, Omer Levy, and Samuel
  Bowman. 2018.
\newblock \href {https://doi.org/10.18653/v1/W18-5446} {{GLUE}: A multi-task
  benchmark and analysis platform for natural language understanding}.
\newblock In \emph{Proceedings of the 2018 {EMNLP} Workshop {B}lackbox{NLP}:
  Analyzing and Interpreting Neural Networks for {NLP}}, pages 353--355,
  Brussels, Belgium. Association for Computational Linguistics.

\bibitem[{Warstadt et~al.(2023{\natexlab{a}})Warstadt, Choshen, Mueller,
  Williams, Wilcox, and Zhuang}]{warstadt2023papers}
Alex Warstadt, Leshem Choshen, Aaron Mueller, Adina Williams, Ethan Wilcox, and
  Chengxu Zhuang. 2023{\natexlab{a}}.
\newblock Call for papers -- the babylm challenge: Sample-efficient pretraining
  on a developmentally plausible corpus.
\newblock \emph{Computing Research Repository}, arXiv:2301.11796.

\bibitem[{Warstadt et~al.(2023{\natexlab{b}})Warstadt, Mueller, Choshen,
  Wilcox, Zhuang, Ciro, Mosquera, Williams, Paranjabe, Linzen, and
  Cotterell}]{warstadt-et-al-2023-babylm}
Alex Warstadt, Aaron Mueller, Leshem Choshen, Ethan~Gotlieb Wilcox, Chengxu
  Zhuang, Juan Ciro, Rafael Mosquera, Adina Williams, Bhargavi Paranjabe, Tal
  Linzen, and Ryan Cotterell. 2023{\natexlab{b}}.
\newblock Findings of the 2023 {B}aby{LM} {C}hallenge: {S}ample-efficient
  pretraining on developmentally plausible corpora.
\newblock In \emph{Proceedings of the 2023 {B}aby{LM} {C}hallenge}. Association
  for Computational Linguistics (ACL).

\bibitem[{Warstadt et~al.(2020{\natexlab{a}})Warstadt, Parrish, Liu, Mohananey,
  Peng, Wang, and Bowman}]{warstadt-etal-2020-blimp-benchmark}
Alex Warstadt, Alicia Parrish, Haokun Liu, Anhad Mohananey, Wei Peng, Sheng-Fu
  Wang, and Samuel~R. Bowman. 2020{\natexlab{a}}.
\newblock \href {https://doi.org/10.1162/tacl_a_00321} {{BL}i{MP}: The
  benchmark of linguistic minimal pairs for {E}nglish}.
\newblock \emph{Transactions of the Association for Computational Linguistics},
  8:377--392.

\bibitem[{Warstadt et~al.(2019)Warstadt, Singh, and
  Bowman}]{warstadt-etal-2019-neural}
Alex Warstadt, Amanpreet Singh, and Samuel~R. Bowman. 2019.
\newblock \href {https://doi.org/10.1162/tacl_a_00290} {Neural network
  acceptability judgments}.
\newblock \emph{Transactions of the Association for Computational Linguistics},
  7:625--641.

\bibitem[{Warstadt et~al.(2020{\natexlab{b}})Warstadt, Zhang, Li, Liu, and
  Bowman}]{warstadt-etal-2020-learning}
Alex Warstadt, Yian Zhang, Xiaocheng Li, Haokun Liu, and Samuel~R. Bowman.
  2020{\natexlab{b}}.
\newblock \href {https://doi.org/10.18653/v1/2020.emnlp-main.16} {Learning
  which features matter: {R}o{BERT}a acquires a preference for linguistic
  generalizations (eventually)}.
\newblock In \emph{Proceedings of the 2020 Conference on Empirical Methods in
  Natural Language Processing (EMNLP)}, pages 217--235, Online. Association for
  Computational Linguistics.

\bibitem[{Williams et~al.(2018)Williams, Nangia, and
  Bowman}]{williams-etal-2018-broad}
Adina Williams, Nikita Nangia, and Samuel Bowman. 2018.
\newblock \href {https://doi.org/10.18653/v1/N18-1101} {A broad-coverage
  challenge corpus for sentence understanding through inference}.
\newblock In \emph{Proceedings of the 2018 Conference of the North {A}merican
  Chapter of the Association for Computational Linguistics: Human Language
  Technologies, Volume 1 (Long Papers)}, pages 1112--1122, New Orleans,
  Louisiana. Association for Computational Linguistics.

\bibitem[{Wilson(2006)}]{pmid21702842}
C.~Wilson. 2006.
\newblock {{L}earning phonology with substantive bias: an experimental and
  computational study of velar palatalization}.
\newblock \emph{Cogn Sci}, 30(5):945--982.

\bibitem[{Wu et~al.(2016)Wu, Schuster, Chen, Le, Norouzi, Macherey, Krikun,
  Cao, Gao, Macherey, Klingner, Shah, Johnson, Liu, Kaiser, Gouws, Kato, Kudo,
  Kazawa, Stevens, Kurian, Patil, Wang, Young, Smith, Riesa, Rudnick, Vinyals,
  Corrado, Hughes, and Dean}]{https://doi.org/10.48550/arxiv.1609.08144}
Yonghui Wu, Mike Schuster, Zhifeng Chen, Quoc~V. Le, Mohammad Norouzi, Wolfgang
  Macherey, Maxim Krikun, Yuan Cao, Qin Gao, Klaus Macherey, Jeff Klingner,
  Apurva Shah, Melvin Johnson, Xiaobing Liu, Łukasz Kaiser, Stephan Gouws,
  Yoshikiyo Kato, Taku Kudo, Hideto Kazawa, Keith Stevens, George Kurian,
  Nishant Patil, Wei Wang, Cliff Young, Jason Smith, Jason Riesa, Alex Rudnick,
  Oriol Vinyals, Greg Corrado, Macduff Hughes, and Jeffrey Dean. 2016.
\newblock \href {https://doi.org/10.48550/ARXIV.1609.08144} {Google's neural
  machine translation system: Bridging the gap between human and machine
  translation}.

\bibitem[{Zbontar et~al.(2021)Zbontar, Jing, Misra, LeCun, and
  Deny}]{pmlr-v139-zbontar21a}
Jure Zbontar, Li~Jing, Ishan Misra, Yann LeCun, and Stephane Deny. 2021.
\newblock \href {https://proceedings.mlr.press/v139/zbontar21a.html} {Barlow
  twins: Self-supervised learning via redundancy reduction}.
\newblock In \emph{Proceedings of the 38th International Conference on Machine
  Learning}, volume 139 of \emph{Proceedings of Machine Learning Research},
  pages 12310--12320. PMLR.

\bibitem[{Zhang et~al.(2022)Zhang, Roller, Goyal, Artetxe, Chen, Chen, Dewan,
  Diab, Li, Lin, Mihaylov, Ott, Shleifer, Shuster, Simig, Koura, Sridhar, Wang,
  and Zettlemoyer}]{zhang2022opt}
Susan Zhang, Stephen Roller, Naman Goyal, Mikel Artetxe, Moya Chen, Shuohui
  Chen, Christopher Dewan, Mona Diab, Xian Li, Xi~Victoria Lin, Todor Mihaylov,
  Myle Ott, Sam Shleifer, Kurt Shuster, Daniel Simig, Punit~Singh Koura, Anjali
  Sridhar, Tianlu Wang, and Luke Zettlemoyer. 2022.
\newblock \href {http://arxiv.org/abs/2205.01068} {Opt: Open pre-trained
  transformer language models}.

\bibitem[{Zhang et~al.(2021)Zhang, Warstadt, Li, and
  Bowman}]{zhang-etal-2021-need}
Yian Zhang, Alex Warstadt, Xiaocheng Li, and Samuel~R. Bowman. 2021.
\newblock \href {https://doi.org/10.18653/v1/2021.acl-long.90} {When do you
  need billions of words of pretraining data?}
\newblock In \emph{Proceedings of the 59th Annual Meeting of the Association
  for Computational Linguistics and the 11th International Joint Conference on
  Natural Language Processing (Volume 1: Long Papers)}, pages 1112--1125,
  Online. Association for Computational Linguistics.

\end{thebibliography}
\bibliographystyle{acl_natbib}

\clearpage
\onecolumn
\appendix

\begin{figure}[!t]
    \includegraphics[width=\textwidth]{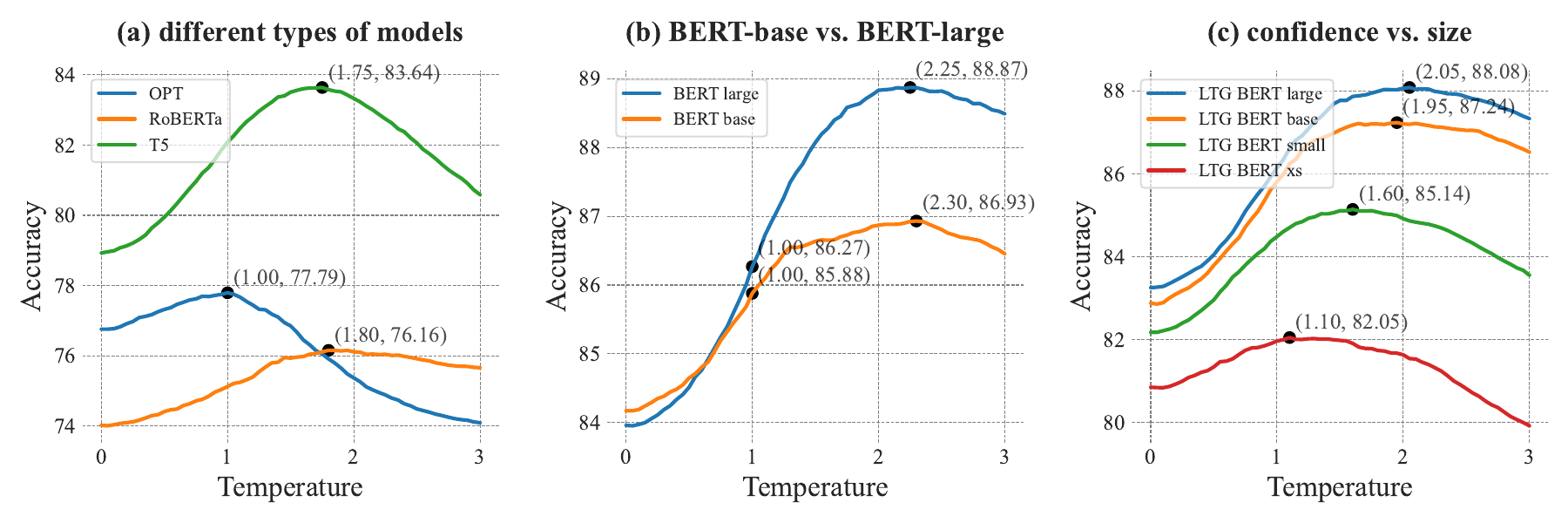}
    \caption{These plots show the BLiMP `confidence profiles' of several language models -- the influence of temperature scaling on the average BLiMP accuracy. \textbf{(a)} Models trained by different training objectives show different confidence profiles, judging their linguistic knowledge from BLiMP accuracy can be misleading. Here, we compare the three baseline from the BabyLM challenge trained on the \textsc{strict} track.
    \textbf{(b)} The linguistic knowledge of BERT\textsubscript{base} and BERT\textsubscript{large} appears comparable when judging from performance at temperature 1, but the potential of the larger model is much greater. \textbf{(c)} We train four sizes of LTG-BERT on the \textsc{strict} track and plot their confidence profiles. Larger models tend to be more confident and, therefore, measuring them at temperature 1 is more misleading.}
    \label{fig:blimp-1}
\end{figure}

\section{The Effect of temperature scaling on BLiMP}
\label{app:blimp}

\begin{figure}[!t]
    \includegraphics[width=\textwidth]{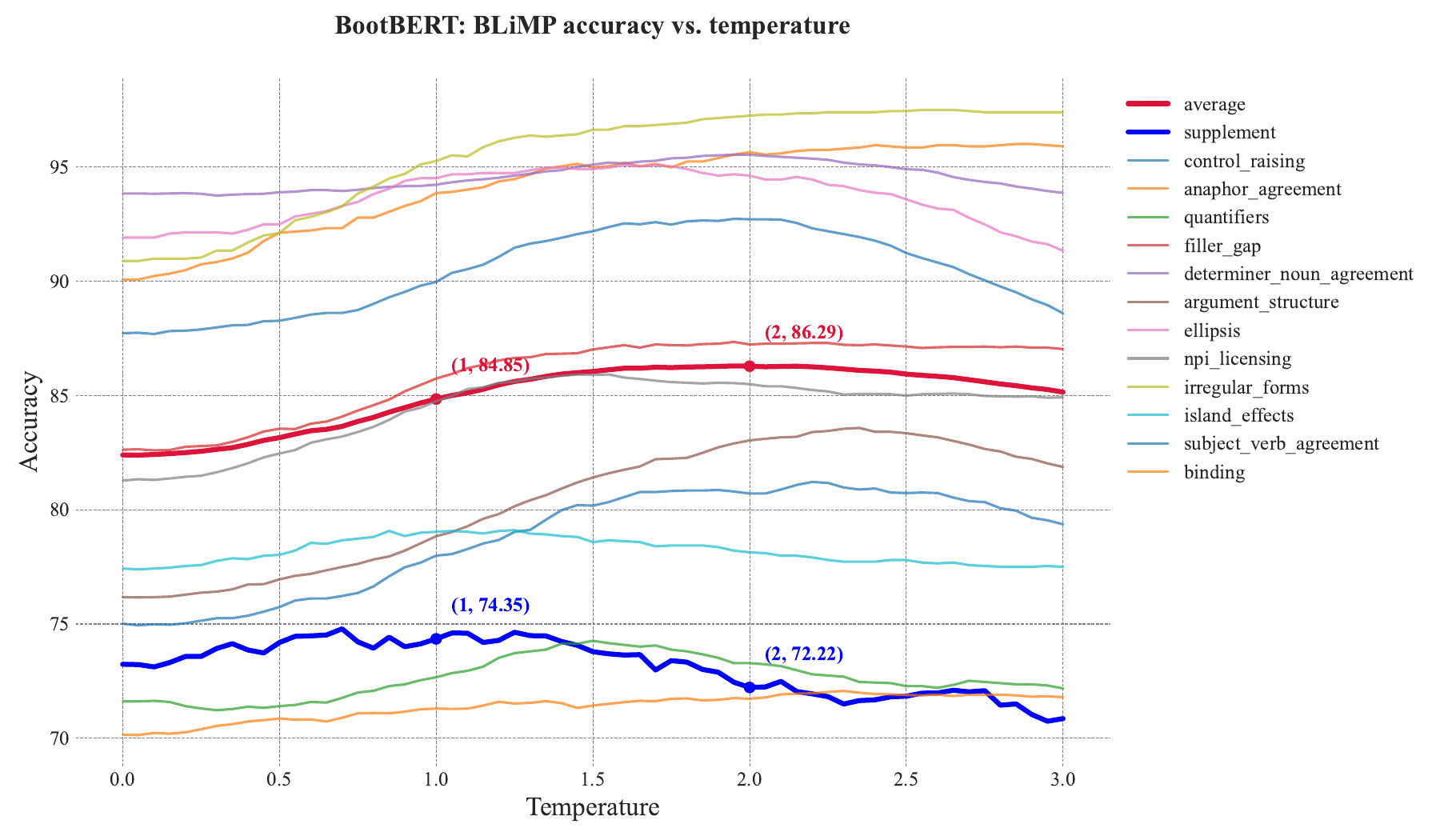}
    \caption{The confidence profile of our proposed BootBERT\textsubscript{base} model pretrained on the \textsc{strict} track. Apart from the average BLiMP accuracy (in red) and the average BLiMP supplement accuracy (in blue), this plot shows fine-grained BLiMP accuracies on all subtasks.}
    \label{fig:blimp-2}
\end{figure}

Our preliminary experiments with calibrating language models via temperature scaling \citep{guo2017calibration} revealed that the BLiMP scores are hugely dependent on the scalar temperature parameter -- when these are calculated with the standard method by \citep{salazar-etal-2020-masked}. This single temperature value can increase the accuracy on some BLiMP subtasks by more than 10\% (\cref{fig:blimp-2}), which challenges the usage of BLiMP as an appropriate evaluation tool. It is especially problematic when comparing different types of language models (\cref{fig:blimp-1}{a}) and different sizes of language models (\cref{fig:blimp-1}{b,c}).

\paragraph{Background.} To better understand this problem, this section describes how are the BLiMP scores traditionally computed for masked language models. These models can estimate $P(\bm{s}_t \vert \bm{s}_{\setminus t})$ -- the likelihood of a token $s_t$ given its bidirectional context $\bm{s}_{\setminus t} = \left(\bm{s}_i \vert i \neq t\right)$. This probability distribution $P$ is given by a softmax transformation of the output logits $z$, where $\tau$ is temperature:

\begin{equation*}
    P_i = \frac{\text{exp}(\nicefrac{z_i}{\tau})}{\sum_k{\text{exp}(\nicefrac{z_k}{\tau})}}.
\end{equation*}

\noindent
Large temperature yields more even distribution and low temperature gives more `peaky' distribution.

\newcite{salazar-etal-2020-masked} proposed to use these probability estimates (with $\tau = 1$) to infer a \textit{score} for each BLiMP sentence, with a higher \textit{score} corresponding to a more likely sentence. Then, the BLiMP accuracy measures how many times is the score of a grammatically correct sentence greater than the score of an incorrect sentence. Specifically, we use the \textit{pseudo-log-likelihood score} (PPL) by \newcite{wang-cho-2019-bert}. The PPL score of a sentence $s$ is defined as:
    $$\textrm{PLL}(\bm{s})=\sum_{t=1}^{N}{\textrm{log}P(\bm{s}_t \vert \bm{s}_{\setminus t})}.$$

\paragraph{Proposed solution.} BLiMP should measure the linguistic knowledge of language models and we believe that this metric should be independent of the prediction confidence of these models. Formally speaking, the BLiMP score should be invariant to temperature scaling. Therefore, we propose to use the maximal average accuracy across all possible temperature values -- instead of simply using the average accuracy at temperature equal to 1. As apparent from \cref{fig:blimp-1}b, such formulation can better reflect the difference of linguistic knowledge found in BERT\textsubscript{base} and BERT\textsubscript{large}. There, the accuracy measured at temperature 1 is at odds with other measures that show substantially better linguistic knowledge of BERT\textsubscript{large} \citep{devlin-etal-2019-bert, tenney2018what, ettinger-2020-bert}.

Note that our approach bears only a negligible compute cost because the temperature modification is done ex-post, i.e., it does not require any additional passes through the language model.

Using one temperature for all subtasks does not account for the severe difference between the accuracy scores on these tasks (\cref{fig:blimp-2}), but it is a simple solution that also allows us to evaluate models on a held-out set, such as the BLiMP supplement. We believe that a scoring function that is (i) unified, (ii) invariant to temperature and (iii) fair to all subtasks, is an interesting future work.

\vspace{1em}

\section{Data preprocessing}
\label{app:data-preprocess}

The pretraining datasets for the \textsc{strict} and \textsc{strict-small} tracks are a mix 10 different corpora, as shown in \cref{tab:data}. We applied light preprocessing and normalization to these corpora in order to cast them into a unified format. In particularly, we applied these modifications:

\begin{itemize}
    \item \textbf{CHILDES}: We capitalize the first letter of each line, normalize punctuation with whitespaces (essentially detokenization) and put every line between double quotes (as directed speech).
    \item \textbf{British National Corpus}: Capitalization, normalization and double quotes.
    \item \textbf{Children's Book Test}: This corpus contains some remnants of the Penn Tree format where, for example, \texttt{-LRB-} and \texttt{-RRB-} tokens are used instead of `(' and `)'. We normalize all unnatural symbols and whitespaces.
    \item \textbf{Children's Stories Text Corpus}: We try to conserve the formatting with a special \texttt{[TAB]} symbol and apply whitespace normalization.
    \item \textbf{Standardized Project Gutenberg Corpus}: The text file is aligned into blocks by inserting a newline symbol after at most 70 characters, which ruins the sentence structure. We restore the original paragraphs by removing these additional newline symbols and apply whitespace normalization.
    \item \textbf{OpenSubtitles}: Some lines arbitrarily start with a dash symbol, which we remove. Then whitespace normalization is applied and every line is cast a directed speech with double quotes.
    \item \textbf{QED}: This corpus contains some incorrectly parsed HTML symbols, which we tried to clean up with some simple heuristics. The whitespace normalization is applied and every line is cast as directed speech with double quotes.
    \item \textbf{Wikipedia}: This dataset also needed to be cleaned of incorrectly parsed Wikipedia tags and hyperlinks. Whitespace normalization is applied.
    \item \textbf{Simple Wikipedia}: Heuristic HTML clean-up and whitespace normalization.
    \item \textbf{Switchboard}: The same as OpenSubtitles: removed leading dashes, whitespaces normalization and added double quotes.
\end{itemize}

\noindent
Note that the preprocessed corpora and the preprocessing scripts are released alongside the training scripts. 

\begin{table*}[h!]
    \centering
    \resizebox{\textwidth}{!}{%
    \begin{tabular}{@{}llrrr@{}}
    \toprule
    & & \multicolumn{2}{c}{\textbf{\# Words}} & \\\cmidrule(lr){3-4}
    Dataset & Domain & \textsc{strict-small} & \textsc{strict} & Proportion \\
    \midrule
    CHILDES \citep{macwhinney2000childes} & Child-directed speech & 0.44M & 4.21M & 5\% \\
    British National Corpus (BNC),\textsuperscript{1} dialogue portion & Dialogue & 0.86M & 8.16M & 8\% \\
    Children's Book Test \citep{hill-2016-cbt} & Children's books & 0.57M & 5.55M & 6\% \\
    Children's Stories Text Corpus\textsuperscript{2} & Children's books & 0.34M & 3.22M & 3\% \\
    Standardized Project Gutenberg Corpus \citep{gerlach-2018-gutenberg} & Written English & 0.99M & 9.46M & 10\% \\
    OpenSubtitles \citep{lison-tiedemann-2016-opensubtitles2016} & Movie subtitles & 3.09M & 31.28M & 31\% \\
    QCRI Educational Domain Corpus (QED; \citealp{abdelali-etal-2014-qed}) & Educational video subtitles & 1.04M & 10.24M & 11\% \\
    Wikipedia\textsuperscript{3} & Wikipedia (English) & 0.99M & 10.08M & 10\% \\
    Simple Wikipedia\textsuperscript{4} & Wikipedia (Simple English) & 1.52M & 14.66M & 15\% \\
    Switchboard Dialog Act Corpus \citep{Stolcke-etal:2000} & Dialogue & 0.12M & 1.18M & 1\% \\
    \midrule
    \emph{Total} & -- & 9.96M & 98.04M & 100\% \\
    \bottomrule
    \end{tabular}}
    \caption{The contents of datasets for the the \textsc{strict} and \textsc{strict-small} tracks; the table is taken from \newcite{warstadt-et-al-2023-babylm}. \textsuperscript{1}\footnotesize\url{http://www.natcorp.ox.ac.uk}\ \ \ \textsuperscript{2}\url{https://www.kaggle.com/datasets/edenbd/children-stories-text-corpus}\ \ \ \textsuperscript{3}\url{https://dumps.wikimedia.org/enwiki/20221220/}\ \ \ \textsuperscript{4}\url{https://dumps.wikimedia.org/simplewiki/20221201/}}
    \label{tab:data}
\end{table*}

\vspace{1em}

\section{Implementation details}
\label{app:implementation-details}

In order to reduce training time, pre-training is parallelized over multiple GPUs with the global batch size of 4\,096. The number of GPUs used depends on the size of pre-trained language models, ranging from 32 to 128 AMD Instinct MI250X GPUs, each with 64GB memory. The amount of training steps is 62\,500, reducing the training budget of the original BERT model by 50\%. Unlike the BERT and LTG-BERT training recipe, we use the same sequence length, 256, throughout the whole training. This decision is necessary for keeping a reasonable exponential moving average of the parameters (it could be corrupted when switching to a longer sequence length in the middle of training).

The implementation of latent bootstrapping mainly follows I-JEPA \citep{assran2023self}. We also adopt their usage of a linearly increasing schedule of the EMA decay hyperparameter $\tau$ and a cosine schedule of weight decay.

The hyperparameters for pretraining are given in \cref{tab:hyperparams}. \cref{tab:glue-hyperparams} shows the finetuning hyperparameters.

\vspace{1em}

\section{Finegrained MSGS scores}
\label{app:msgs}

This section shows the full score distribution over all MSGS subtasks, including the control subtasks. This gives a better view on the behavior of different language models than the aggragated scores in \cref{fig:msgs} and \cref{tab:metrics}.

\begin{figure}[h!]
\begin{subfigure}{\textwidth}
   \includegraphics[trim={0 1cm 0 0},clip,width=1\textwidth]{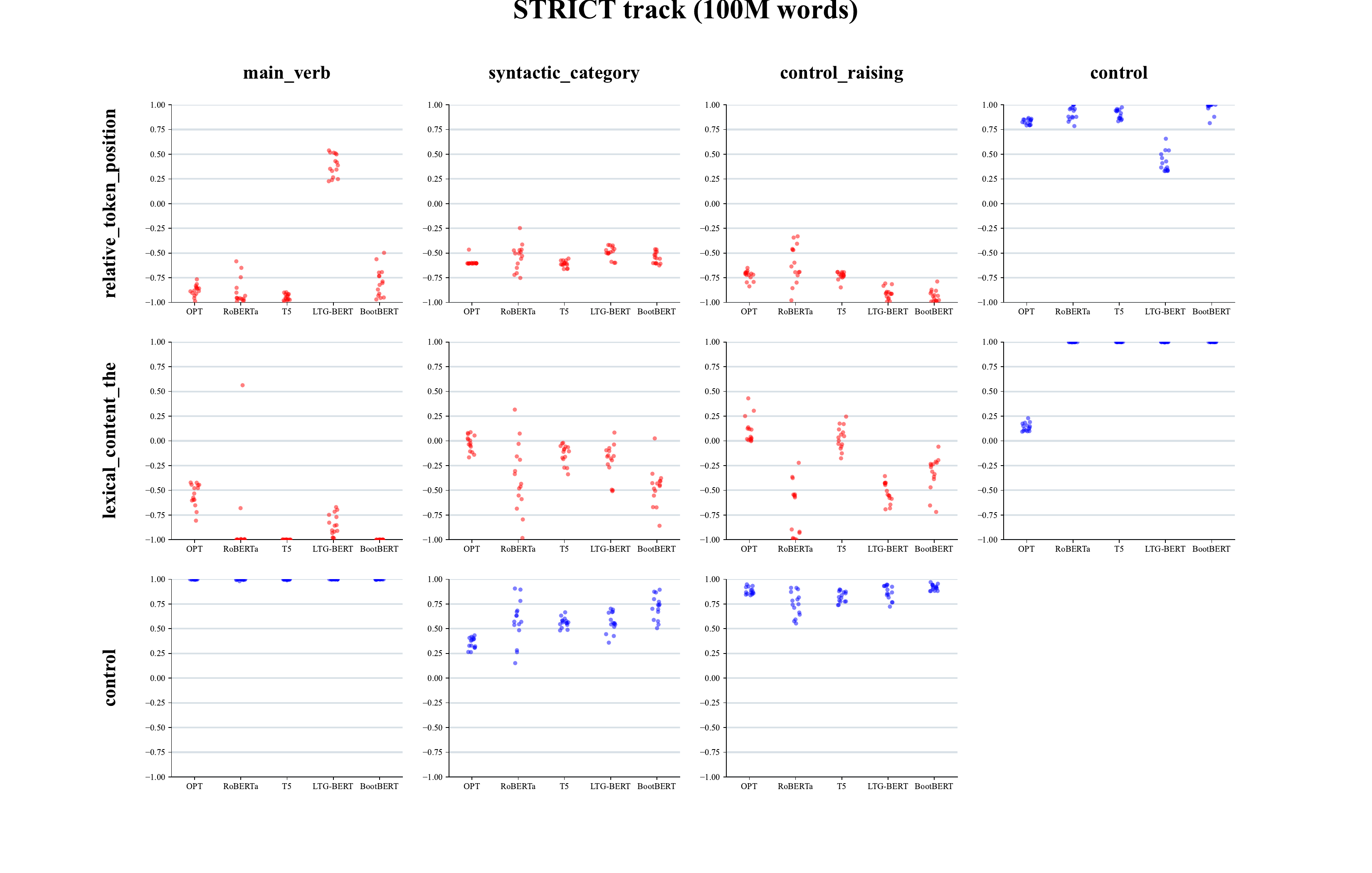}
   \label{fig:Ng1} 
\end{subfigure}

\begin{subfigure}{\textwidth}
   \includegraphics[trim={0 3cm 0 0},clip,width=1\textwidth]{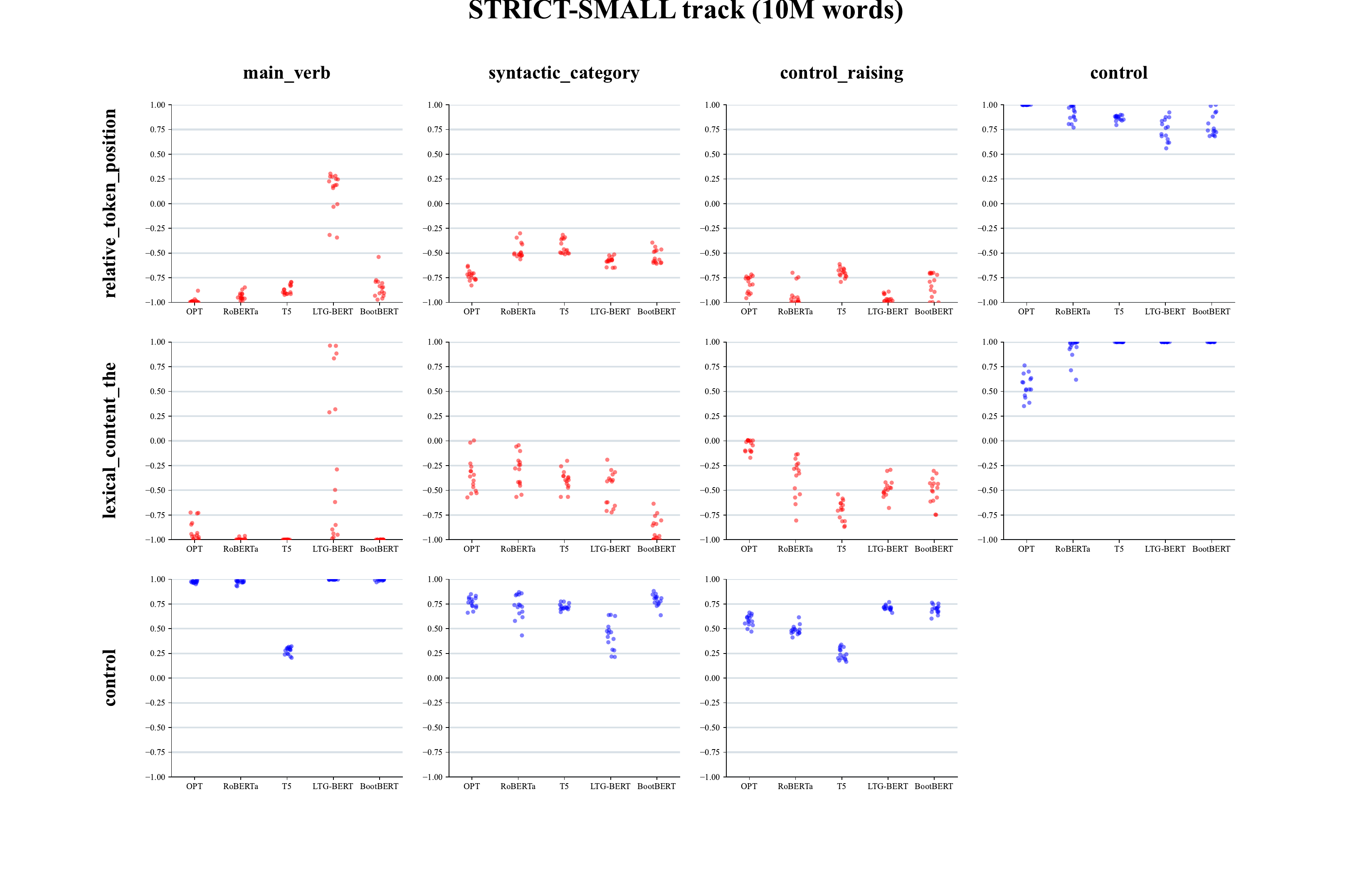}
   \label{fig:Ng2}
\end{subfigure}

\caption{The MSGS linguistic bias scores of the control tasks (in blue) and non-control disambiguated tasks (in red). Values close to 1 indicate preference of linguistic explanations (columns) while values close to -1 indicate preference of surface explanations.}
\end{figure}

\vspace{1em}

\section{The official BabyLM results from DynaBench}
\label{app:dynabench}

This section shows the official results for the BabyLM challenge as published on the DynaBench website.\footnote{\url{https://dynabench.org/babylm} (22 October, 2023)} We show the top 9 submissions (the official ones delivered on time) for the \textsc{strict} and \textsc{strict-small} tracks with the aggregated scores in \cref{tab:dynabench}.

\begin{table}[h!]
\resizebox{\textwidth}{!}{%
\begin{tabular}{@{\hspace{1em}}l@{}rrr@{\hspace{2em}}r@{}}
\multicolumn{3}{@{}l}{\textbf{\textsc{strict} track (100M words)}} \\[0.5em]
\toprule
\textbf{Model} & \textbf{BLiMP} & \textbf{GLUE} & \textbf{MSGS} & \textbf{Average}\\\midrule
BootBERT & \textsuperscript{ \#2 }82.2 & \textsuperscript{ \#1 }\textbf{78.5} & \textsuperscript{ \#3 }27.7 & \textsuperscript{ \#3 }70.2 \\[1em]
ELC-BERT & \textbf{82.8} & 78.3 & 47.2 & \textbf{74.3} \\
Contextualizer & 79.0 & 72.9 & \textbf{58.0} & 73.0 \\
MSLM & 76.2 & 73.5 & 21.4 & 64.4 \\
Bad babies & 77.0 & 67.2 & 23.4 & 63.4 \\
CogMemLM & 72.8 & 72.2 & -0.1 & 58.0 \\
Pre-training LLMs & 71.6 & 69.8 & -3.8 & 56.0 \\
BabyStories & 73.9 & 59.1 & 0.2 & 54.7 \\
AB-RoBERTa & 68.3 & 64.1 & -11.8 & 51.0 \\
\bottomrule
\end{tabular}
\quad \quad
\begin{tabular}{@{\hspace{1em}}l@{}rrr@{\hspace{2em}}r@{}}
\multicolumn{3}{@{}l}{\textbf{\textsc{strict-small} track (10M words)}} \\[0.5em]
\toprule
\textbf{Model} & \textbf{BLiMP} & \textbf{GLUE} & \textbf{MSGS} & \textbf{Average}\\\midrule
BootBERT & \textsuperscript{ \#1 }\textbf{75.9} & \textsuperscript{ \#2 }71.7 & \textsuperscript{ \#9 }-9.7 & \textsuperscript{ \#7 }57.5 \\[1em]
ELC-BERT & 75.8 & \textbf{73.7} & \textbf{29.4} &	\textbf{65.9} \\
MLSM & 72.4	& 70.6 & 17.2 &	60.8 \\
Contextualizer & 74.3 & 69.6 & 12.7 & 60.5 \\
Baby Llama & 69.8 & 67.6 & 24.7 & 60.1 \\
Too Much Information & 75.7 & 70.9 & 3.9 & 59.9 \\
McGill & 72.4 & 69.3 & 5.2 & 58.0 \\
CLIMB & 71.8 & 65.6 & 9.7 & 57.5 \\
William's college GPT2 & 70.9 & 64.8 & 9.9 & 56.9 \\
\bottomrule
\end{tabular}%
}
\caption{\label{tab:dynabench}
The DynaBench scores of the BabyLM challenge \citep{warstadt2023papers}, the table shows the top 9 submissions in the \textsc{strict} and \textsc{strict-small} tracks. Higher scores are better, the best results in each evaluation suite are boldfaced. 
}
\end{table}


\section{BabyLM subset of (Super)GLUE tasks}
\label{app:glue}

The BabyLM challenge involves slightly modified GLUE and SuperGLUE benchmarks. It uses only a subset of the subtasks, the datasets are filtered so that they do not contain out-of-vocabulary words, and it sometimes use non-standard metrics. We list all subtasks and their metrics below:

\begin{itemize}\itemsep0em 
    \item \textbf{Boolean Questions} \citep[BoolQ;][]{clark-etal-2019-boolq}, a yes/no Q/A dataset evaluated with accuracy.
    \item \textbf{Corpus of Linguistic Acceptability} \citep[CoLA;][]{warstadt-etal-2019-neural} evaluated with accuracy (originally evaluated with the Matthews correlation coefficient \citep[MCC;][]{MATTHEWS1975442}).
    \item \textbf{The Multi-Genre Natural Language Inference Corpus} \citep[MNLI;][]{williams-etal-2018-broad}. Its development set consists of two parts: \textit{matched}, sampled from the same data source as the training set, and \textit{mismatched}, which is sampled from a different domain. Both parts are evaluated with accuracy.
    \item \textbf{The Microsoft Research Paraphrase Corpus} \citep[MRPC;][]{dolan-brockett-2005-automatically}, evaluated with both F\textsubscript{1}-score (originally also evaluated with accuracy).
    \item \textbf{Multi-Sentence Reading Comprehension} \citep[MultiRC;][]{khashabi-etal-2018-looking}, a multiple choice question answering dataset, evaluated with accuracy (originally evaluated with the exact match accuracy (EM) and F\textsubscript{1}-score (over all answer options)).
    \item \textbf{Question-answering Natural Language Inference} (QNLI) constructed from the Stanford Question Answering Dataset \citep[SQuAD;][]{rajpurkar-etal-2016-squad}, evaluated with accuracy.
    \item \textbf{The Quora Question Pairs} (QQP),\footnote{\url{https://quoradata.quora.com/First-Quora-Dataset-Release-Question-Pairs}} evaluated with F\textsubscript{1}-score (originally evaluated with accuracy).
    \item \textbf{The Stanford Sentiment Treebank} \citep[SST-2;][]{socher-etal-2013-recursive}, evaluated with accuracy.
    \item \textbf{The Recognizing Textual Entailment datasets} \citep[RTE;][]{10.1007/11736790_9, rte2, giampiccolo-etal-2007-third, Bentivogli09thefifth}, evaluated with accuracy.
    \item \textbf{Winograd Schema Challenge} \citep[WSC;][]{10.5555/3031843.3031909} evaluated with accuracy.
\end{itemize}

\vspace{1em}

\begin{table*}[h!]
\centering
\tiny
\begin{tabular}{@{}lcc@{}}
\toprule
\textbf{Hyperparameter} & \textbf{BootBERT\textsubscript{small}} & \textbf{BootBERT\textsubscript{base}} \\ \midrule
Number of parameters    & 30\,395\,776 & 127\,744\,768 \\
Number of layers        & 12 & 12        \\
Hidden size             & 384 & 768         \\
FF intermediate size    & 1\,024 & 2\,048    \\
Vocabulary size         & 4\,096 & 16\,384          \\
Attention heads         & 6 & 12           \\
$\beta$ parameter       & 0.1 & 0.1 \\
Encoder hidden dropout                 & 0.1   & 0.1        \\
Encoder attention dropout       & 0.1   & 0.1          \\
Decoder hidden dropout                 & 0.0   & 0.0        \\
Decoder attention dropout       & 0.0   & 0.0          \\
Training steps          & 62\,500 & 62\,500     \\
Batch size              & 4\,096 & 4\,096      \\
Sequence length         & 256 & 256        \\
Warmup steps            & 1\,000 & 1\,000         \\
Initial learning rate   & 0.007 & 0.005          \\
Final learning rate     & 0.0007 & 0.0005          \\
Learning rate scheduler & cosine & cosine        \\
Initial weight decay    & 0.04 & 0.02          \\
Final weight decay      & 0.4 & 0.2 \\
Weight decay scheduler  & cosine & cosine \\
Initial EMA decay $\tau$       & 0.996 & 0.996 \\
Final EMA decay $\tau$        & 1.0 & 1.0 \\
EMA decay scheduler     & linear & linear \\ 
Layer norm $\epsilon$   & 1e-7 & 1e-7          \\
Optimizer               & LAMB & LAMB         \\
LAMB $\epsilon$         & 1e-6 & 1e-6          \\
LAMB $\beta_1$          & 0.9  & 0.9         \\
LAMB $\beta_2$          & 0.98   & 0.98       \\
Gradient clipping       & 2.0  & 2.0         \\ \bottomrule
\end{tabular} %
\caption{Pre-training hyperparameters for the small-sized BootBERT (trained on \textsc{strict-small} and for the base-sized BootBERT (trained on the \textsc{strict} track).}
\label{tab:hyperparams}
\end{table*}

\begin{table*}
\tiny
\centering
\begin{tabular}{@{}lccc@{}}
\toprule
\multirow{3}{*}{\textbf{Hyperparameter}} & \textbf{BoolQ, MNLI} & \multirow{3}{*}{\textbf{CoLA, RTE, WSC}} & \multirow{3}{*}{\textbf{MSGS}} \\
    & \textbf{MRPC, MultiRC, QNLI} & & \\
    & \textbf{QQP, SST-2} & & \\ \midrule

Batch size             & 32 & 16 & 16  \\
Number of epochs       & 10 & 10 & 5   \\
Dropout                & 0.1& 0.1 & 0.1 \\
Warmup steps           & 10\%  & 10\%                     & 10\% \\
Peak learning rate    & 3e-5     & 3e-5                     & \{1e-5, 2e-5, 3e-5\}      \\
Learning rate decay  & linear    & linear                   & linear  \\
Weight decay           & 0.01   & 0.01                     & 0.01 \\
Optimizer             & AdamW        & AdamW                    & AdamW \\
Adam $\epsilon$     & 1e-6    & 1e-6                     & 1e-6 \\
Adam $\beta_1$       & 0.9        & 0.9                      & 0.9    \\
Adam $\beta_2$     & 0.999        & 0.999                    & 0.999  \\ \bottomrule
\end{tabular}
\caption{Hyperparameters for fine-tuning the GLUE, SuperGLUE task and MSGS tasks. We use the same hyperparameters for all models, not performing any per-model hyperparameter search. These values are adopted from LTG-BERT \citep{samuel-etal-2023-trained} and MSGS \citep{warstadt-etal-2020-learning}. For all models, we measure the statistics over 5 random seeds: 1234, 2345, 3456, 4567 an 5678.}
\label{tab:glue-hyperparams}
\end{table*}

\end{document}